\documentclass[final,5p,times,twocolumn]{elsarticle}

\usepackage{amsmath,amsfonts,amssymb,amsthm}

\usepackage{threeparttable}
\usepackage{multirow}
\usepackage{lineno}
\usepackage{fixltx2e}
\MakeRobust{\overrightarrow}
\usepackage{xcolor}
\usepackage{textcomp}
\usepackage{setspace}
\usepackage{hyperref}
\modulolinenumbers[5]

\journal{Journal of Biomedical Informatics}

\begin{document}

\begin{frontmatter}



\title{Explainable Automated Coding of Clinical Notes using Hierarchical Label-wise Attention Networks and Label Embedding Initialisation}

\author[uoe,hdr]{Hang Dong}
\author[uoe,hdr]{V\'{i}ctor Su\'{a}rez-Paniagua}
\author[ccbs,hdr]{William Whiteley}
\author[ucl,hdr]{Honghan Wu}

\address[uoe]{Centre for Medical Informatics, Usher Institute of Population Health Sciences and Informatics, University of Edinburgh, Edinburgh, United Kingdom}
\address[ccbs]{Centre for Clinical Brain Sciences, University of Edinburgh}
\address[ucl]{Institute of Health Informatics, University College London, London, United Kingdom}
\address[hdr]{Health Data Research UK, London, United Kingdom}

\begin{abstract}
\textit{Background:} Diagnostic or procedural coding of clinical notes aims to derive a coded summary of disease-related information about patients. Such coding is usually done manually in hospitals but could potentially be automated to improve the efficiency and accuracy of medical coding. Recent studies on deep learning for automated medical coding achieved promising performances. However, the explainability of these models is usually poor, preventing them to be used confidently in supporting clinical practice. Another limitation is that these models mostly assume independence among labels, ignoring the complex correlations among medical codes which can potentially be exploited to improve the performance.
\vskip 0.03in
\noindent\textit{Methods:} To address the issues of model explainability and label correlations, we propose a Hierarchical Label-wise Attention Network (HLAN), which aimed to interpret the model by quantifying importance (as attention weights) of words and sentences related to each of the labels. Secondly, we propose to enhance the major deep learning models with a label embedding (LE) initialisation approach, which learns a dense, continuous vector representation and then injects the representation into the final layers and the label-wise attention layers in the models. We evaluated the methods using three settings on the MIMIC-III discharge summaries: full codes, top-50 codes, and the UK NHS (National Health Service) COVID-19 (Coronavirus disease 2019) shielding codes. Experiments were conducted to compare the HLAN model and label embedding initialisation to the state-of-the-art neural network based methods, including variants of Convolutional Neural Networks (CNNs) and Recurrent Neural Networks (RNNs).
\vskip 0.03in
\noindent\textit{Results:} HLAN achieved the best Micro-level AUC and $F_1$ on the top-50 code prediction, 91.9\% and 64.1\%, respectively; and comparable results on the NHS COVID-19 shielding code prediction to other models: around 97\% Micro-level AUC. More importantly, in the analysis of model explanations, by highlighting the most salient words and sentences for each label, HLAN showed more meaningful and comprehensive model interpretation compared to the CNN-based models and its downgraded baselines, HAN and HA-GRU. Label embedding (LE) initialisation significantly boosted the previous state-of-the-art model, CNN with attention mechanisms, on the full code prediction to 52.5\% Micro-level $F_1$. The analysis of the layers initialised with label embeddings further explains the effect of this initialisation approach. The source code of the implementation and the results are openly available at \url{https://github.com/acadTags/Explainable-Automated-Medical-Coding}.
\vskip 0.03in
\noindent\textit{Conclusion:} We draw the conclusion from the evaluation results and analyses. First, with hierarchical label-wise attention mechanisms, HLAN can provide better or comparable results for automated coding to the state-of-the-art, CNN-based models. Second, HLAN can provide more comprehensive explanations for each label by highlighting key words and sentences in the discharge summaries, compared to the $n$-grams in the CNN-based models and the downgraded baselines, HAN and HA-GRU. Third, the performance of deep learning based multi-label classification for automated coding can be consistently boosted by initialising label embeddings that captures the correlations among labels. We further discuss the advantages and drawbacks of the overall method regarding its potential to be deployed to a hospital and suggest areas for future studies.
\end{abstract}


\end{frontmatter}

\section{Introduction}
\label{intro}
Diagnostic or procedural coding of medical free-text documents (e.g. discharge summaries) aims to derive a coded summary of disease-related information about patients, for clinical care, audit, and research. In hospitals, such coding is usually done manually, requiring much cognitive human effort, but could potentially be automated. An automated program could efficiently take a clinical note as input and then output medical codes from existing classification systems, e.g. ICD (International Classification of Diseases). This could facilitate coding professionals provide more accurate results.

This clinical task is technically challenging, due to (i) the explainability required to process long documents, in average about 2000 tokens in a discharge summary in MIMIC-III \cite{baumel2018multi}, and thus pose a ``needle-in-a-haystack'' issue to locate the key words and sentences relevant to each code; (ii) the complex label correlations in the multi-label setting, in average about 16 different ICD-9 (the Ninth Revision) codes per discharge summary in the MIMIC-III dataset \cite{johnson_mimic-iii_2016}, which inherently exhibit the complex relations among codes; and (iii) a large set of codes when using all the codes as candidates for prediction, e.g. around 13k unique codes in ICD-9 and many times further in ICD-10 \cite{Cartwright2013} and ICD-11 \cite{Stewart2018}.

Automated medical coding has been studied for more than a decade. Early studies mostly use systems based on rules, grammar, and string matching, as reviewed in \cite{stanfill_systematic_2010}. Recent studies adapt deep learning based document classification methods, which commonly formalise the task as a \textit{multi-label classification} problem \cite{karimi-etal-2017-automatic,mullenbach-etal-2018-explainable,baumel2018multi}. Typically, they use variations of Recurrent Neural Networks (RNNs) and Convolutional Neural Networks (CNNs) to derive a continuous representation of clinical notes matched to the high-dimensional coding space. However, few studies have tackled the above challenges above regarding explainability and label correlations.

Explainability (or interpretability, used interchangeably in this paper) is a key requirement for models applied to the clinical domain, particularly regarding the ethical aspect and to build medical professionals' trust in machine learning models \cite{geis2019ethics,goodman2017european}. Also, to facilitate the work of coding professionals, a desired automated coding system should be able to highlight the most essential part of a long clinical note to support the assignment of medical codes. To address this, models based on CNNs can be adapted to highlight $n$-gram information to support the explanation, as in \cite{mullenbach-etal-2018-explainable}. Solely the $n$-grams, however, may not be enough to provide accurate interpretation reflecting the document structure.

In this work, we propose to highlight the most essential words and sentences in a document for automated medical coding, inspired and adapted from Hierarchical Attention Networks (HAN) \cite{yang2016} and the recent model, Hierarchical Attention bi-directional Gated Re-current Units (HA-GRU) \cite{baumel2018multi}. With attention mechanisms, HAN can highlight the salient words and sentences related to the overall prediction. However, HAN could not generate a specific interpretation for each label. HA-GRU \cite{baumel2018multi} can provide a \emph{sentence}-level explanation for each label, but still could not specify the most essential \emph{words} leading to the decision of each code. We present a novel model, Hierarchical Label-wise Attention Network (HLAN), which has label-wise word-level and sentence-level attention mechanisms, so as to provide a richer explainability of the model.

We formally evaluated HLAN along with HAN, HA-GRU, and CNN-based neural network approaches for automated medical coding. With better or comparative coding performance in various data settings,  HLAN can further generate more comprehensive explanations through key sentences and words for each label, as indicated from the analysis on model explainability. The analysis of the false positive predictions also shows that the explanation based on the hierarchical label-wise attention mechanisms in HLAN can serve as a reference for medical professionals and engineers to make reasonable coding decisions and system iterations even when the model seems to predict erroneously.

Apart from model interpretability, another issue not thoroughly studied in deep learning based multi-label classification is label correlation. Medical codes are related and can be predicted together, for example, the code 486 (ICD 9 for Pneumonia) commonly appeared for over 1.5k times (out of about 53k documents) with the code 518.81 (Acute respiratory failure) in the MIMIC-III dataset. Such co-occurrences are under-lied by the clinical, biomedical, and biological associations among different diseases. Deep learning for multi-label classification represents the label space with orthogonal vectors: each label as a one-hot vector and each label set as a multi-hot representation \cite{Nam2014,mullenbach-etal-2018-explainable}. This, however, assumes independence among labels.

We propose an effective label embedding initialisation approach to tackle the label correlation problem. We encode the label correlation using pre-trained label embeddings from the label sets in the training data, derived from the coding practice. Then the label embeddings are used to initialise the weights in the final hidden layer and label-wise attention layers. The idea is that the linear projection can automatically leverage the label similarity encoded in the continuous label embedding space. This approach shows consistent and significant improvement, while not requiring hyper-parameter tuning or further computational complexity.

We evaluate our approach with three specific datasets based on the openly available, MIMIC-III database \cite{johnson_mimic-iii_2016}, containing clinical notes in the critical care sector in the US. The first two datasets, full code and top-50 code predictions, are the same as in the work \cite{mullenbach-etal-2018-explainable} for comparison. The third dataset was created to simulate the task of identifying high-risk patients for shielding during the COVID-19 (Coronavirus disease 2019) pandemic by predicting the ICD-9 codes matched to the codes used in the UK NHS (National Health Service) patient shielding identification method\footnote{\url{https://digital.nhs.uk/coronavirus/shielded-patient-list/methodology}}.

Thus, the contribution of the paper includes:
\begin{itemize}
    \item A novel, Hierarchical Label-wise Attention Network (HLAN) for automated medical coding. The proposed HLAN model provides an explanation in the form of attention weights on both the word level and the sentence level for the prediction of each medical code.
    \item An effective label embedding (LE) initialisation approach to enhance the performance of various deep learning models for multi-label classification. Analysis of the LE initialised layers shows the efficacy to leverage label correlations for medical coding.
    \item A formal comparison of the main deep learning based methods for automated coding. Experiments on three datasets based on the MIMIC-III discharge summaries, i.e. full code prediction, top-50 code prediction, and the NHS COVID-19 shielding-related code prediction, show the advantage of the proposed method over the state-of-the-art methods (CNNs, Bi-GRU) and downgraded baselines (HA-GRU, HAN). Label embedding initialisation significantly improved the performance of neural network models in most evaluation settings. An analysis and comparison of the model interpretability demonstrate the most comprehensive explanations from the HLAN model.
\end{itemize}

The rest of the paper is organised as follows. First, we review the related work on automated medical coding with explainability, deep learning methods for multi-label classification, and label correlation in Section \ref{rw}. Then, we present the problem formulation, followed by the proposed model, HLAN, and the idea of LE initialisation in Section \ref{method}. The experiments, including datasets, experimental settings, main and per-label results, analysis and comparison of model explainability, and analysis on the layers initialised with LE, are in Section \ref{experiment}. We finally discuss the advantages and drawbacks of the overall methods in Section \ref{discussion} and summarise the work in Section \ref{conclusion}.

\section{Related Work}
\label{rw}
We will first present the task of automated medical coding with the methods used especially in most recent studies, then introduce in detail the mainstream breakthrough on deep learning-based multi-label classification for the task, and finally review the label correlation issue, particularly relevant to the medical and clinical domain.

\subsection{Automated Medical Coding with Explainability}
\label{rw:medical_coding}
Automated medical coding is the task of transforming medical records, especially the natural language in the clinical notes, into a set of structured, medical codes to facilitate clinical care, audit, and research \cite{stanfill_systematic_2010}. The applied alphanumerical codes in the clinical domain, such as ICD and SNOMED-CT, represent patients' diagnosis, procedures and other information with controlled clinical terminology.

One of the earliest reviews back in 2010 \cite{stanfill_systematic_2010} surveyed 113 studies on coding or classification of clinical notes. Most of the studies applied tools with rule-based, grammar-based, and string matching methods, and they in overall suffered the challenges of reasoning and the lack of method generalisability. The field of automated medical coding has in more recent years been advanced with the open, benchmarking datasets like radiology reports in \cite{pestian2007} and MIMIC-III \cite{johnson_mimic-iii_2016} discharge summaries. With the datasets, \textit{deep learning} based approaches have been proposed and tested, which have generally demonstrated better performance than traditional machine learning methods. The work in \cite{karimi-etal-2017-automatic} compared the deep learning based method, CNN, with several traditional machine learning methods, support vector machine, random forests, and logistic regression, for ICD-9 code prediction (number of ICD-9 codes $|Y|$=38) from 978 radiology reports in \cite{pestian2007}. The result showed comparable or improved results of the deep learning approach to the traditional methods, even without parameter tuning in the CNN model. The work in \cite{mullenbach-etal-2018-explainable} adapted CNN with attention mechanisms and established a state-of-the-art performance in predicting the full set ($|Y|$=8,922) and the top-50 most frequent ICD-9 codes ($|Y|$=50) from MIMIC-III discharge summaries.

A key aspect of clinical applications is their requirement of the \textit{explainability} of models. Users are entitled to a ``right of explanation'' when their data being used for AI algorithms, as potentially regulated by the General Data Protection Regulation (GDPR) \cite{goodman2017european}. For clinical applications, e.g. radiology, the Joint European and North American Multisociety Statement raises great ethical concern on AI algorithms regarding explainability, i.e. ``the ability to explain what happened when the model made a decision, in terms that a person understands'' \cite[p.~438]{geis2019ethics}. While deep learning achieves better results in general, the approach is inherently less transparent than traditional methods due to its extremely complex networks of non-linear activation.

Few studies explored the explainability of deep learning models for automated medical coding. A representative work is the study \cite{mullenbach-etal-2018-explainable}, which compared the ability of different models to highlight $n$-grams along with the models' ICD-9 code prediction. A manual evaluation showed that the CNN model with attention mechanisms can generate more meaningful $n$-grams relevant to the labels \cite{mullenbach-etal-2018-explainable}. The study \cite{baumel2018multi} proposed a Hierarchical Attention bi-directional Gated Recurrent Unit (HA-GRU) to produce a sentence-level explanation for each code, instead of $n$-gram-level explanation. In this work, we propose an approach with enhanced interpretability, from both the label-wise word-level and the sentence-level attention weights, to support automated coding.

\subsection{Deep Learning-based Multi-label Classification with Attention Mechanisms}
\label{rw:dl}
Automated medical coding is mainly formulated as a multi-label classification problem \cite{zhang2014,Tsoumakas2010,mullenbach-etal-2018-explainable,baumel2018multi}, where each object (e.g. clinical note) is associated with a set of labels (e.g. diagnosis or procedure ICD codes) instead of a single label in binary or multi-class classification.

\textit{Deep learning} has become the main approach for multi-label document classification \cite{Nam2014,dong2020} in recent years. The advantage of multi-label deep learning models lies in their straightforward problem formulation and strong approximation power on large datasets, resulting in better performance over traditional machine learning approaches, as compared in \cite{zhang2006,dong2020,karimi-etal-2017-automatic}. For automated coding, some of the notable neural network models adapted for multi-label classification are variations of CNNs \cite{karimi-etal-2017-automatic,mullenbach-etal-2018-explainable} and RNNs \cite{baumel2018multi} with attention mechanisms. Pre-trained models with multi-head self-attention blocks (e.g. BERT, Bi-directional Encoder Representations from Transformers) \cite{devlin-etal-2019-bert}, while substantially improved many NLP tasks, so far still are under-performing for automated coding with the MIMIC-III discharge summaries \cite{chalkidis2020empirical,Chen2020icd9bert}.

The idea of the above mentioned \textit{attention mechanism} is a key, recent advancement in deep learning for NLP, originated from machine translation to align (or attend to) words in the source sentence in one language to predict each of the target words in another language \cite{bahdanau2014}. This inspires to jointly learn to represent the important words and sentences while classifying a document in HAN \cite{yang2016}, thus also enables to \textit{explain} the inner working of deep learning models. HAN was adapted to a multi-label classification setting to classify socially shared texts in \cite{dong2020} and for automated medical coding \cite{baumel2018multi}. Founded on the studies above, our approach provides a richer label-wise attention mechanism at both the word and the sentence level for automated medical coding.

\subsection{Label Correlation}
\label{rw:label_correlation}
In multi-label classification, labels are potentially correlated to each other. As the example in Section \ref{intro}, the medical codes of ``Pneumonia'' and ``Acute Respiratory Failure'' tend to appear together in the MIMIC-III discharge summaries. In automated medical coding, the number of unique code $|Y|$ is large ($|Y|=8,922$ in the MIMIC-III dataset) and further the possible label relations (e.g. the number of pairwise combinations is near to $|Y|^2$). Such correlations among the labels represent additional knowledge that could be exploited to improve performance \cite{Gibaja2015}.

This issue of \textit{label correlation} (or ``label dependence'') remains an ongoing challenge \cite{Gibaja2015} in multi-label classification, especially with deep learning models. Deep learning for multi-label classification mostly represents the label space with orthogonal vectors: each label as a one-hot vector and each label set as a multi-hot representation, in general domains \cite{Nam2014} and clinical domains \cite{mullenbach-etal-2018-explainable,baumel2018multi}. Combined with the sigmoid activation and binary cross-entropy loss, this overall approach, effectively, assumes independence among labels.

One recent approach to address the problem is through weight initialisation \cite{kurata-etal-2016-improved,baker-korhonen-2017-initializing}: initialising higher weights for dedicated neurons (each encoding a co-occurrence relation among labels) in the final hidden layer. The approach showed performance improvement, however, it is not computationally efficient to assign each neuron in the final hidden layer to represent one of the massive (even the pairwise) patterns of label relations. An alternative method is through regularisation in \cite{dong2020} to enforce the output layer of the neural network to satisfy constraints on label relations. This requires to further tune the hyper-parameters of the regularisers so that a relatively marginal improvement (0.5-1.5\% example-based $F_1$ on scientific paper abstracts and questions in social Q\&A platforms) could be achieved. In this study, we further propose a novel effective weight initialisation approach to tackle the label correlation problem, by initialising pre-trained dense label embeddings instead of the sparse co-occurrence representations.

\section{Proposed Method}
\label{method}
We formalise automated medical coding from clinical notes as a multi-label text classification problem \cite{Nam2014}. With deep learning, multi-label classification mainly contains two, integrated parts, (i) a neural document encoder, representing documents into a continuous representation, and (ii) a prediction layer, matching the document space to the label space. We present the problem formalisation and the deep learning based multi-label classification in Section \ref{problem_formulation}. Then, regarding the neural document encoder, we propose the hierarchical label-wise attention network in Section \ref{hlan}, followed by the idea of label embedding initialisation in the prediction layer in Section \ref{le}.

\subsection{Problem Formulation with Deep Learning Models}
\label{problem_formulation}
Formally multi-label classification can be defined as follows. Suppose $X$ denoting the collection of textual sequences (e.g. clinical notes), and $Y = \{y_1,y_2,...,y_{|Y|}\}$ denotes the full set of labels (i.e. ICD codes) of size $|Y|$. Each instance $x_d \in X$ is a word sequence of a document, where $d$ is the document index. Each $x_d \in X$ is associated with a label set $Y_d \subseteq Y$. Each label set $Y_d$ can be represented as a $|Y|$-dimensional \emph{multi-hot} vector, $\overrightarrow{Y_d} = [y_{d1},y_{d2},...,y_{d|Y|}]$ and $y_{dl}\in\{0,1\}$, where a value of $1$ indicates that the $l$th label $y_{l}$ has been used to annotate (is relevant to) the $d$th instance, and $0$ indicates irrelevance. The task is to learn a complex function $f:X \rightarrow Y$ based on a training set $D = \{x_d,\overrightarrow{Y_d}|d\in[1,m]\}$, where $m$ is the number of instances in the training set.

Neural document encoders in deep learning models (e.g. CNN, RNN, and BERT, as review in Section \ref{rw:dl}) represent each word sequence $x$ as a continuous vector $h$, with matrix projection and non-linear activation. The representation $h$ is projected to the label space and turned into $p_{dl} \in (0,1)$ with the sigmoid function ($ \sigma(x) =  \frac{\mathrm{1} }{\mathrm{1} + e^x }  $), as defined in Equation \ref{projection} below, where the weight $w_l$ (a row vector in $W$) and the bias $b$ are parameters to be learned during the training process. The obtained $p_{dl}$ is the probability of the label (e.g. ICD code) $y_l$ being related to the document (e.g. discharge summary) $d$.
\begin{equation}
\label{projection}
    p_{dl} = \sigma (w_lh + b) \text{, or collectively as }  p_{d} = \sigma (Wh + b)
\end{equation}

The loss function is commonly the binary cross-entropy loss \cite{Nam2014} as defined in Equation \ref{CE}, which measures the sum of negative log-likelihood of the predictions $p_{dl}$ of the actual labels. A large deviation between $\overrightarrow{Y_{dl}}$ and $p_{dl}$ will cause a greater value in the $L_{CE}$ and thus will be penalised during training.
\begin{equation}\label{CE}
  L_{CE} = -\sum_{d} \sum_{l} (\overrightarrow{Y_{dl}}\log (p_{dl})  + (1-\overrightarrow{Y_{dl}})\log (1-p_{dl}))
\end{equation}

For inference, a calibration threshold $\mathit{Th}$ (default as 0.5) is set to assign the label to the document when $p_{dl} > \mathit{Th}$.

\subsection{Hierarchical Label-wise Attention Network}
\label{hlan}
\begin{figure*}[h]
  \center
  \includegraphics[width=0.7\textwidth]{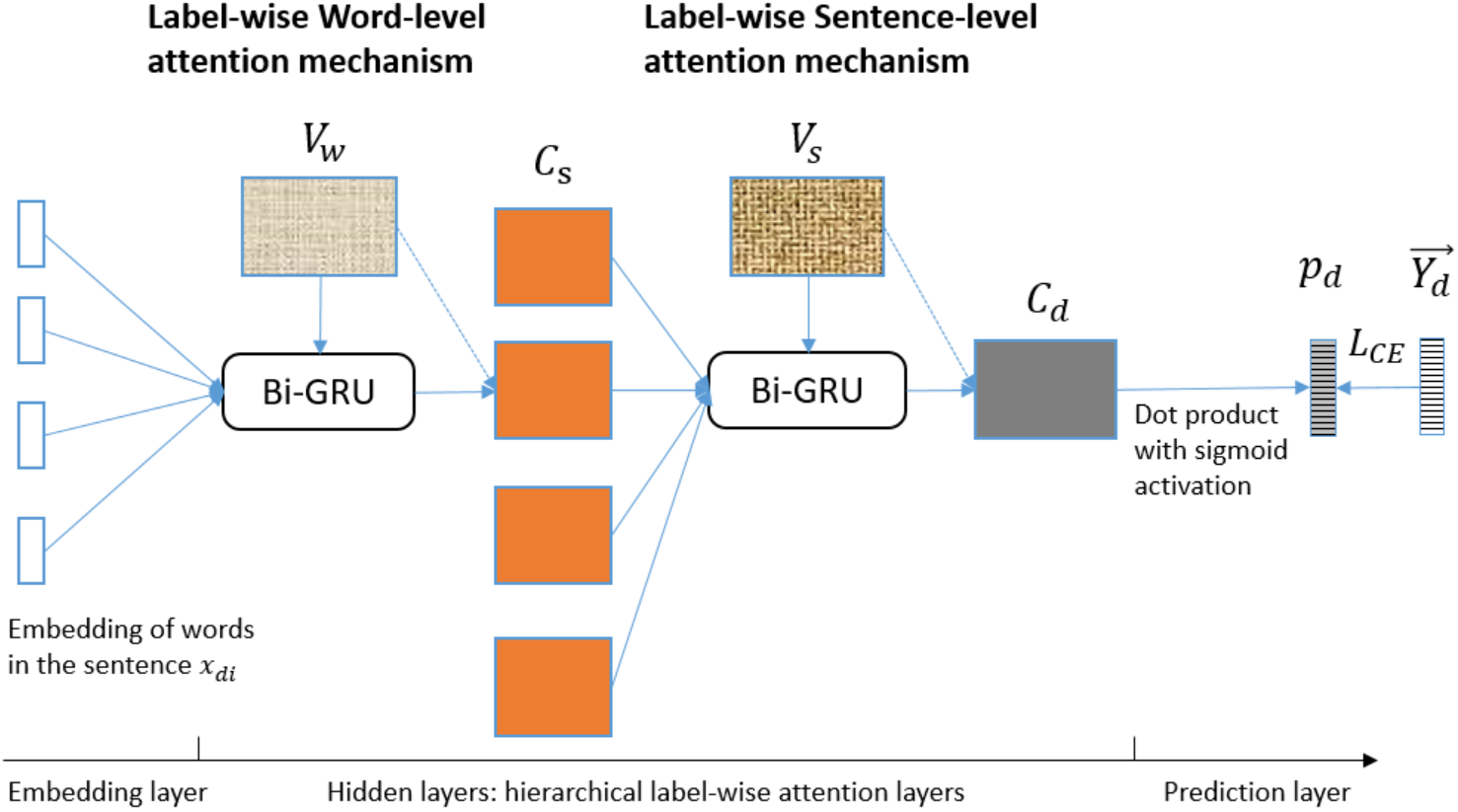}
  \caption{Hierarchical Label-wise Attention Network (HLAN)}\label{hlan_architecture}
\end{figure*}

Following the framework above, the neural document encoder in HLAN (as illustrated in Figure \ref{hlan_architecture}) takes into input the word sequence $x_d = \{x_{d1},x_{d2},...,x_{dn}\}$, where $x_{di}$ denotes the sequence of tokens in the $i$th of all $n$ \textit{sentences}, and output the document representation. The distinction to HAN \cite{yang2016} is that HLAN represents the same document differently at both the word-level and the sentence-level regarding different labels. HLAN extends the contextual vectors in HAN to the label-wise contextual matrices, $V_w$ and $V_s$. The document representation also becomes a matrix, $C_d$, where each row (corresponding to each label) has the same dimensionality as $h$.

As shown in Figure \ref{hlan_architecture}, the model consists of an embedding layer, hidden layers (hierarchical label-wise attention layers), and a prediction (or projection) layer. First, the embedding layer transforms the one-hot input representation $u_{di}$ of each token in the sequence of the $i$th sentence $x_{di}$ into a low-dimensional continuous vector, $e_{di}=W_{e}u_{di}$, where we used the neural word embedding algorithm, Word2vec \cite{mikolov2013distributed}, to pre-train $W_{e}$ for its efficiency.

Second, we applied the Gated Recurrent Unit (GRU) \cite{cho2014}, a type of RNN unit, to capture long-term dependencies in the clinical narrative. An RNN unit ``reads'' each token in the sequence one by one, every time producing a new hidden state $h^{(t)}$, corresponding to the token at time $t$. Different from the vanilla RNN unit, GRU additionally considers the previous tokens by using a reset gate $r^{(t)}$ and an update gate $z^{(t)}$. This allows to model the dependencies among tokens in long sequences. A GRU can be formally defined as in the Equations \ref{gru} below, where $\overrightarrow{h}$ denotes the hidden states through forward processing, $\sigma$ is a non-linear activation function (e.g. sigmoid function), $W_{hr},W_{hz},W_{h\tilde{h}} \in \mathbb{R}^{d_h \times d_h}$ are weights, and $b_r, b_z \in \mathbb{R}^{d_h}$ represent bias terms. A bi-directional adaptation was applied by concatenating the hidden states at each time after read the sequence both forwardly ($\rightarrow$) and backwardly ($\leftarrow$) to form a more comprehensive representation, $h^{(t)} = [\overrightarrow{h}^{(t)};\overleftarrow{h}^{(t)}] \in \mathbb{R}^{2d_h}$. This sub-architecture is generally known as Bi-GRU \cite{cho2014}.
\begin{equation}
\label{gru}
\begin{split}
  r^{(t)} & = \sigma (W_{er}e^{(t)} + W_{hr}\overrightarrow{h}^{(t-1)} + b_r) \\
  z^{(t)} & = \sigma (W_{ez}e^{(t)} + W_{hz}\overrightarrow{h}^{(t-1)} + b_z) \\
  \tilde{h}^{(t)} & = \tanh(W_{e\tilde{h}}e^{(t)} + W_{h\tilde{h}}(r^{(t)} \circ \overrightarrow{h}^{(t-1)})) \\
  \overrightarrow{h}^{(t)} & = (1-z^{(t)}) \circ \overrightarrow{h}^{(t-1)} + z^{(t)} \circ \tilde{h}^{(t)}
\end{split}
\end{equation}
For simplicity, we use the function $h = \text{Bi-GRU}(e,\Theta)$ to denote the whole process (with bi-directional concatenation of hidden states) above. Instead of applying one single Bi-GRU layer to represent the whole document, we applied a word-level Bi-GRU to represent each sentence and then a sentence-level one to represent the whole document, as illustrated in Figure \ref{hlan_architecture}. This captures the hierarchical structure of the document and relieves the burden of having a too lengthy sequence for each GRU \cite{baumel2018multi} (e.g. from the original sequence length 2500 in the MIMIC-III discharge summaries to only 100 on the word level and 25 on the sentence level).

A common way is to represent the whole sequence as the concatenated hidden state $h^{(t)}$ of the last time $t$. This representation tends to emphasise the ending elements (i.e. words or sentences) and does not discriminate between the elements in a sequence. In fact, the key information for medical coding is contained in a well selected part of the lengthy discharge summary. We therefore use an attention mechanism to learn a weighted average of the hidden states to form a final representation as in \cite{yang2016,bahdanau2014}. The attention scores are based on an alignment (or a similarity computation) of each hidden representation in a sequence to a context vector. The context vector is usually \textit{shared} for all labels as in \cite{yang2016,dong2020}, whereas in medical coding, it is essential to interpret the amount of attention paid regarding a \textit{specific} medical code to the clinical note.
\begin{equation}
\label{word-att}
\begin{split}
  h^{(i)} & = \text{Bi-GRU}(e, \Theta_w)\\
  v^{(i)} & = \tanh (W_{w}h^{(i)} + b_{w}) \\
  \alpha_{wl}^{(i)} & = \frac{\exp (V_{wl} \bullet v^{(i)})}{\sum_{o \in [1,\mathrm{n_t}]} \exp(V_{wl} \bullet v^{(o)})} \\
  C_{sl} & = \sum_{i \in [1,\mathrm{n_t}]} \alpha_{wl}^{(i)} h^{(i)}
\end{split}
\end{equation}
Thus, the adapted, \textit{label-wise word-level attention mechanism} is defined in Equations \ref{word-att} above. The context matrix for the word-level attention mechanism is denoted as $V_w \in \mathbb{R}^{|Y| \times d_w}$, where each row $V_{wl}$ (of attention layer size $d_w$) is the context vector corresponding to the label $y_l$. The attention score $\alpha_{wl}^{(i)}$ for the label $y_l$ is calculated as a softmax function of the dot product similarity between the vector representation $v^{(i)}$ (transformed from the $i$th hidden state $h^{(i)}$ with a feed-forward layer) and the context vector $V_{wl}$ for the same label. $\mathrm{n_t}$ denotes the number of tokens in a sentence. The sentence representation $C_{sl}$, as a row vector in $C_s \in \mathbb{R}^{|Y| \times 2d_h}$, for the label $y_l$, is computed as the weighted average of all the hidden state vectors $h^{(i)}$.

In a similar way, we can compute the \textit{label-wise sentence-level attention mechanism} as defined in Equations \ref{sent-att}, which encodes each row $C_{sl}$ in the sentence representations $C_{s}$ to a label-wise sentence representation $S^{(r)}_l$, to be non-linearly transformed to $U^{(r)}_l$ and aligned to the corresponding row $V_{sl}$ in sentence-level contextual matrix $V_{s} \in \mathbb{R}^{|Y| \times d_s}$, and outputs the sentence-level attention scores $\alpha_{sl}$ (for a label $y_l$) and the document representation matrix $C_d \in \mathbb{R}^{|Y| \times 4d_h}$. To note that the dimensionality of $S^{r}_l$ and thus $C_{dl}$ are further doubled to $4d_h$ through the Bi-GRU process.
\begin{equation}
\label{sent-att}
\begin{split}
  S^{(r)}_l & = \text{Bi-GRU}(C_{sl}, \Theta_S)\\
  U^{(r)}_l & = \tanh (W_{S}S^{(r)}_l + b_{S}) \\
  \alpha_{sl}^{(r)} & = \frac{\exp (V_{sl} \bullet U^{(r)}_l)}{\sum_{q \in [1,n]} \exp(V_{sl} \bullet U^{(q)}_l)} \\
  C_{dl} & = \sum_{r \in [1,n]} \alpha_{sl}^{(r)} S^{(r)}_l
\end{split}
\end{equation}

Then, we use a label-wise, dot product projection with logistic sigmoid activation to model the probability of each label to each document, as defined in Equation \ref{projection_label_wise}, adapted from Equation \ref{projection}. The parameters in $w_l$ are row vectors in the projection matrix $W$.
\begin{equation}\label{projection_label_wise}
  p_{dl} = \sigma (w_lC_{dl} + b_l)
\end{equation}
We finally optimise the binary cross-entropy loss function in Equation \ref{CE} with $L_2$ regularisation using the Adam optimiser~\cite{kingma2014}.

\subsection{Label Embedding Initialisation}
\label{le}
For automated medical coding, the diagnostic and procedural codes (or labels) have complex semantic relations, and can potentially be leveraged to improve prediction. Clinically, these code relations represent the correlation among diseases and medical procedures from the medical coding practice.

As we reviewed in Section \ref{rw:label_correlation}, previous studies on weight initialisation to address the label correlation issue mostly focus on a co-occurrence based representation of labels \cite{kurata-etal-2016-improved,baker-korhonen-2017-initializing}. Both studies dedicate a neuron in the final hidden layer to initialise one single co-occurrence pattern. There are, however, very limited neurons to be assigned to initialise the massive number of label relations, especially for the large label size in automated coding.

Instead of encoding the sparse co-occurrence patterns of labels, we learn low-dimensional, dense, label embeddings. For two correlated labels $y_j$ and $y_k$, e.g. 486 (Pneumonia) and 518.81 (Acute respiratory failure), one would expect that the prediction of one label has an impact on the other label for some clinical notes, i.e. $p_{dj}$ is correlated or has a similar value to $p_{dk}$. To achieve this, according to Equations \ref{projection} or \ref{projection_label_wise}, we propose to initialise their corresponding weights $w_j$ and $w_k$ (corresponding to the labels $y_j$ and $y_k$) in $W$ with a label representation $E$ which reflects the actual label correlation (e.g. similarity between $y_j$ and $y_k$) in a continuous space.

\begin{figure*}[t]
  \centering
  \includegraphics[width=0.8\textwidth]{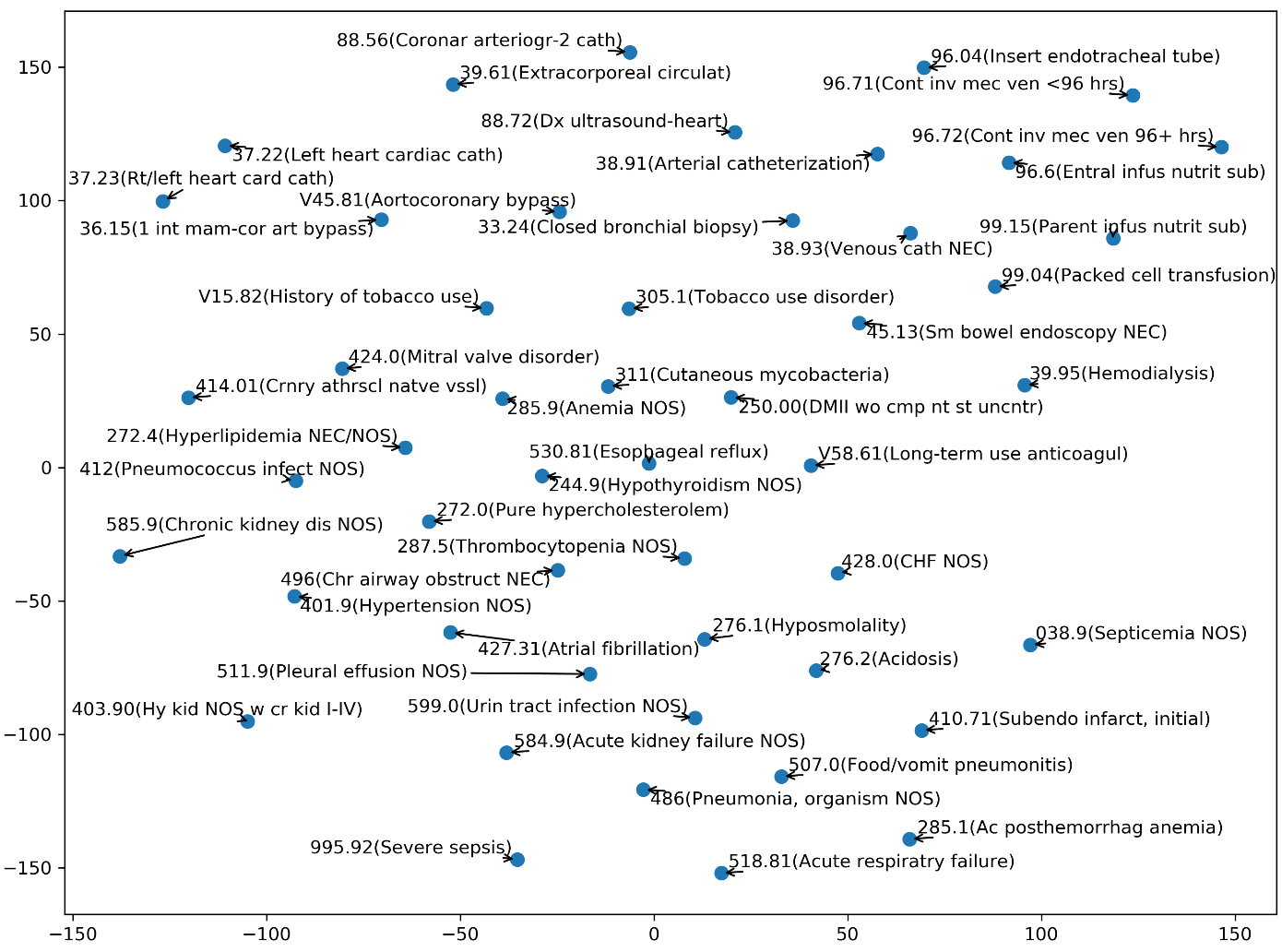}
  \caption{The 2-dimensional T-SNE plot of word2vec Continuous Bag of Words label embeddings of the 50 ICD-9 codes in MIMIC-III-50, trained on the whole training label sets, $\{\protect \overrightarrow{Y_d}|d\in[1,m]\}$, in MIMIC-III.}
  \label{LE-plot}
\end{figure*}

A straightforward idea is thus to initialise the projection matrix $W$ using $E$ as pre-trained label embeddings, e.g. with a neural word embedding algorithm, learned from the label sets in the training data, $\{\overrightarrow{Y_d}|d\in[1,m]\}$. For initialisation, we pre-train the label embeddings $E$ with dimensionality the same as $W$. We used the Continuous Bag of Words algorithm in word2vec \cite{mikolov2013distributed} for its efficiency and its power to represent the correlations of the labels. Figure \ref{LE-plot} shows an intuitive visualisation, for which we used an unsupervised technique, T-SNE (t-distributed Stochastic Neighbor Embedding), to reduce the dimensionality of the learned label embeddings while preserving the local similarity and structure of the labels \cite{maaten2008tsne}. It can be observed that the ICD-9 code learned from the MIMIC-III training label sets can capture the semantic relations that are distinct from the ICD-9 hierarchy. For example, 486 (Pneumonia) and 518.81 (Acute respiratory failure) appear closely on the bottom while they are not under the same parent in the ICD-9 hierarchy.

Besides, the label embedding initialisation can also be applied to the context matrices $V_w$ and $V_s$ (see Figure \ref{hlan_architecture}) in the label-wise attention mechanisms. Taking the word-level attention mechanisms in Equation \ref{word-att} as an example, we can initialise $V_{wl}$ with the pre-trained label embedding $E_l$ for the label $y_l$. This imposes a tendency for context vectors of the correlated labels to align the Bi-GRU encoded token representation $v_i$ in a geometrically similar way. Similarly, we can also initialise the label-wise attention layer in CNN+att \cite{mullenbach-etal-2018-explainable} and in HA-GRU \cite{baumel2018multi}. While the initialised layers are dynamically updated during the training, the tendency that imposed by label embeddings remains for most neural networks; we will empirically demonstrate this in the analysis of initialised layers in Section \ref{le_analysis}.

For automated coding, the approach can be extended by initialising label embeddings with the clinical ontologies and description texts of ICD codes. However, due to the different nature of the knowledge (i.e. embedded label relations), the external sources may bring contradictory label correlations to the ones in the dataset, as also discussed in \cite{song2019}. In this research, we focus on leveraging label relations from the label sets alone as in \cite{kurata-etal-2016-improved,baker-korhonen-2017-initializing}, as it directly reflects the label correlation of the coding practice that generated the dataset, and leave the integration of external knowledge for a future study.

\section{Experiments}
\label{experiment}

We tested HLAN and several strong baseline models, with label embedding initialisation, on three data sets based on the MIMIC-III database. The main results show the comparative results of HLAN to other state-of-the-art models on the datasets and the consistent improvement with label embedding initialisation. More importantly, through an analysis on model interpretability, we also show that HLAN can provide a more comprehensive explanation using the label-wise word and sentence-level attention mechanisms. Analysis of the layers initialised with label embeddings further reveals the effect of the initialisation approach. The source code of our implementation and the results are openly available at \url{https://github.com/acadTags/Explainable-Automated-Medical-Coding}.

\subsection{Datasets}
\label{subsec:data}
We used the benchmark dataset, MIMIC-III (``Medical Information Mart for Intensive Care'') \cite{johnson_mimic-iii_2016}, which contains clinical data from adult patients admitted to the critical care unit in the Beth Israel Deaconess Medical Center in Boston, Massachusetts between 2001 and 2012, to validate our approach. The ICD-9 codes annotated by professionals in the dataset were used as labels. We focused on discharge summaries and followed the preprocessing and data split from \cite{mullenbach-etal-2018-explainable}. The preprocessed full MIMIC-III dataset has 8,922 unique codes as labels assigned to 52,724 discharge summaries, where 47,724 of them (from 36,998 patients) were used for training, 1,632 for validation, and 3,372 for testing. We also used the same top-50 setting (termed as ``MIMIC-III-50'') from \cite{mullenbach-etal-2018-explainable}, which narrows down the labels to the top 50 by their frequencies (codes and their frequencies are available in Table S1 in the supplementary material). This has 8,066 discharge summaries for training, 1,573 for validation, and 1,729 for testing.

We further created a subset of discharge summaries annotated using the COVID-19 shielding related ICD codes. This simulates the application of identifying key patients for shielding during the pandemic. We used the ICD-9 codes matched to the ICD-10 codes selected by the NHS to identify patients with medium or high risks during COVID-19. The considered patients were related to solid organ transplant recipients, people with specific cancers, with severe respiratory conditions, with rare diseases and inborn errors of metabolism, on immunosuppression therapies, or who were pregnant with significant congenital heart disease\footnote{A clearer description of the ``high risk'' category is in \url{https://digital.nhs.uk/coronavirus/shielded-patient-list/methodology/background}}, which is still in active use and under maintenance at the time of writing this paper. While the actual EHR data and the shielded patient list from the NHS are not easy to obtain, the ICD-10 codes are openly available for re-use\footnote{To see the annexe B in \url{https://digital.nhs.uk/coronavirus/shielded-patient-list/methodology/annexes}}. We thus used MIMIC-III to simulate the task of identifying patients for shielding during COVID-19. We selected those appeared at least 50 times in the MIMIC-III dataset, resulting in 20 ICD-9 codes (out of 79 matched codes), available in Table S2 in the supplementary material. After filtering the MIMIC-III dataset with the selected ICD-9 codes, there are 4,574 discharge summaries for training, 153 for validation, and 322 for testing. We name this dataset as ``MIMIC-III-shielding''.

Statistics of the three datasets are in Table \ref{data_stat}. Denoted by \textit{Ave}, the average number of labels per document (or label cardinality) in the training set of MIMIC-III, MIMIC-III-50, and MIMIC-III-shielding are 15.88, 5.69, and 1.08, respectively. While all originated from MIMIC-III database, the three datasets represent different case scenarios in automated medical coding with various scales of data and vocabulary size (``vocab''), number of labels to predict, and the average number of labels per document. While the full MIMIC-III dataset has much more training instances, it is more complex as its number of labels $|Y|$ and vocabularies are significantly greater than MIMIC-III-50 and MIMIC-III-shielding.
\begin{table}[h]
\caption{Statistics of the datasets}
\footnotesize
\center
\label{data_stat}
\begin{tabular}{p{2.4cm}p{0.7cm}p{0.6cm}p{0.5cm}p{0.5cm}p{0.5cm}p{0.5cm}}
\cline{1-7}
Dataset                  & Vocab & Train & Valid & Test & $|Y|$ & \textit{Ave}\\
\cline{1-7}
MIMIC-III-50             & 59,168 & 8,066  & 1,573 & 1,729 & 50 & 5.69\\
MIMIC-III-shielding      & 47,979 & 4,574  & 153 & 322 & 20 & 1.08\\
MIMIC-III                & 140,795 & 47,724  & 1,632 & 3,372 & 8,922 & 15.88\\
\cline{1-7}
\end{tabular}
\end{table}

Figures of ICD-9 code distributions by frequency in the three datasets are available in Figure S1 in the supplementary material, along with the list of the selected codes (and their frequencies) in the MIMIC-III and MIMIC-III-shielding datasets. The statistics show a high imbalanced characteristics of the labels in all three data settings. Most label occurrences are from a few labels and there is a long-tail of labels having very low frequencies. This is most pronounced in the full label setting (``MIMIC-III'') and also presented in the other two datasets.

\subsection{Experiment Settings}

We implemented the proposed Hierarchical Label-wise Attention Network (HLAN) model and the other baselines for comparison:
\begin{enumerate}
    \item CNN, Convolutional Neural Network, which is essentially based on \cite{kim-2014-convolutional} for text classification, and applied in \cite{karimi-etal-2017-automatic,gehrmann2018} for automated medical coding.
    \item CNN+att (or CAML), CNN with a label-wise attention mechanism, proposed in \cite{mullenbach-etal-2018-explainable}.
    \item Bi-GRU, Bi-directional Gated Recurrent Unit \cite{cho2014} for multi-label classification. The document representation is set as the last concatenated hidden state $h^{(t)}$ .
    \item HAN, Hierarchical Attention Network \cite{yang2016}, which can be considered as a downgraded model of HLAN when the attention mechanisms are shared for all labels (see Figure \ref{hlan_architecture}, when $V_w$, $V_s$, and $C_s$, $C_d$ become vectors, same for all labels).
    \item HA-GRU, Hierarchical Attention bi-directional Gated Recurrent Unit, proposed in \cite{baumel2018multi}, which can be considered as a downgraded model of HLAN when the word-level attention mechanism is shared for all labels (see Figure \ref{hlan_architecture}, when $V_w$ and $C_s$ become vectors, same for all labels, while $V_s$ and $C_d$ are the same as in HLAN).
\end{enumerate}

We applied the label embedding initialisation approach (denoted as ``\textbf{+LE}'') to all the models above. We pre-trained the label embeddings $E$ from the label sets in the training data with the word2vec (Continuous Bag of Words with negative sampling) algorithm \cite{mikolov2013distributed}. The label embeddings have the dimension same as the final hidden layer or the label-wise attention layer(s) in each neural network model. We applied the Python Gensim package \cite{rehurek_lrec2010} to train embeddings, by setting the window size as 5 and minimum frequency threshold (``min\_count'') as 0. Label embeddings were normalised to unit length for initialisation. Xavier initialisation \cite{glorot2010} was used for labels not existing in the training data for faster model convergence. We used the same setting to train and initialise the 100-dimension word embeddings $W_e$ from the documents.

The implementations of HLAN and HA-GRU were adapted from our previous implementation\footnote{\url{https://github.com/acadTags/Automated-Social-Annotation/tree/master/2\%20HAN}} of HAN in \cite{dong2020} using the Python Tensorflow \cite{Abadi2016} framework, originated from brightmart's implementation\footnote{\url{https://github.com/brightmart/text_classification}}, all under the MIT license. We adapted HA-GRU with the sigmoid activation and binary cross-entropy as described in Section \ref{problem_formulation}, instead of the softmax activation used in the original paper \cite{baumel2018multi}, for a controlled comparison with other models. For CNN, CNN+att, and Bi-GRU, we adapted the implementation\footnote{\url{https://github.com/jamesmullenbach/caml-mimic}} from \cite{mullenbach-etal-2018-explainable} using the PyTorch framework \cite{NEURIPS2019_9015} with the same parameters for MIMIC-III and MIMIC-50 from \cite{mullenbach-etal-2018-explainable}. For MIMIC-III-shielding, we used the same hyper-parameters as in MIMIC-50. We did not get the results with HA-GRU and HLAN for the MIMIC-III dataset, due to the memory limit caused by the large label size ($|Y|=8,922$), while for MIMIC-III-50 and MIMIC-III-shielding, we obtained the results of all models.

The input token length for the models was padded to 2,500 as in \cite{mullenbach-etal-2018-explainable}. We optimised the precision@$k$ or micro-$F_1$ metrics (defined in Section \ref{metrics}) during the training\footnote{We optimised precision@$k$ for CNN, CNN+att, and Bi-GRU for MIMIC-III and MIMIC-III-50, and micro-$F_1$ for all other models and for the MIMIC-III-shielding dataset.}, according to the implementation in \cite{mullenbach-etal-2018-explainable}.  The batch size for CNN, CNN+att, BiGRU were set as 16 as in \cite{mullenbach-etal-2018-explainable}, for HLAN and HA-GRU as 32, and HAN as 128. For HLAN, HAN, and HA-GRU, we tried both a customised rule-based parsing of real sentences with Spacy\footnote{We parsed sentences using the rule-based pipeline component in Spacy with adding double newlines as another rule to segment sentences, see \url{https://spacy.io/usage/linguistic-features\#sbd}.} and using text chunks of fix length as ``sentences''; for both ways, we set the sentence length as 25 and padded the number of sentences to 100. The dimensions of the final document representation were 512, 500, 50, 400 for Bi-GRU, CNN, CNN+att, and HLAN (also HAN and HA-GRU), respectively. All models were trained using a single GeForce GTX TITAN X server, and the trained HLAN, HA-GRU, and HAN models were further tested using a CPU server (4-core, Intel(R) Xeon(R) Platinum 8259CL CPU @ 2.50GHz). The detailed hyper-parameter settings, containing learning rate, dropout rate, and CNN specific parameters (kernel size and filter size), with the estimated training and testing times, are in Table S3 in the supplementary material.

We also experimented with BERT as the neural document encoder. Due to the GPU memory limit, we tested the normal size of a BERT model, i.e. BioBERT-base \cite{Lee2019BioBert}, which had been further pre-trained with PubMed paper abstracts\footnote{\url{https://pubmed.ncbi.nlm.nih.gov/}} and full texts\footnote{\url{https://www.ncbi.nlm.nih.gov/pmc/}}; we used a sliding window approach to address the token limit issue (512 tokens) in BERT. Our results from the BioBERT-base model were similar to the results in \cite{Chen2020icd9bert}, significantly worse than HLAN and CNN\footnote{We thus do not report the BERT results here but make the implementation details and results available on \url{https://github.com/acadTags/Explainable-Automated-Medical-Coding}.}. We believe further adaptations are necessary for BERT models on automated medical coding and leave the direction for a future study.

\subsection{Evaluation Metrics}
\label{metrics}
For comparison, we applied the same set of label-based metrics as in \cite{mullenbach-etal-2018-explainable} and according to the evaluation of multi-label classification algorithms \cite{Tsoumakas2010,zhang2014}. The chosen metrics include micro- and macro-averaging precision (P), recall (R), $F_1$ score ($F_1$), area under the receiver operating characteristic curve (AUC), an the precision@$k$.

The micro-averaging metrics treat each document-label as a separate prediction, whereas the macro-averaging metrics are an average of the per-label results. Micro- and macro-averaging applies to all the binary evaluation metrics including precision, recall, AUC. For example, the micro- and macro-averaged precision is defined in Equation \ref{mi/macro-prec} below. Recall is calculated in a similar way, but divided by all the true cases ($TP_l+FN_l$), and $F_1$ is then the harmonic mean of the calculated precision and recall, i.e. $F_1 = \frac{2 \times \text{P} \times \text{R}}{\text{P}+\text{R}}$. The AUC is defined by two metrics, the true positive rate (or recall) on the Y axis and false positive rate on the X axis, depicting the tradeoff between the two metrics when varying the calibration threshold $\mathit{Th}$ \cite{fawcett2006}. The overall performance of a classifier (with a set of varied $\mathit{Th}$) can thus be reflected by AUC.
\begin{equation}\label{mi/macro-prec}
  \text{Micro-P} = \frac{\sum_{l=1}^{L}\textit{TP}_l}{\sum_{l=1}^{L}\textit{TP}_l+\textit{FP}_l} \qquad \text{Macro-P} = \frac{1}{|L|}\sum_{l=1}^{L}\frac{\textit{TP}_l}{\textit{TP}_l+\textit{FP}_l}
\end{equation}

In some clinical application or epidemiological studies, only one type of code (either the diagnosis or the procedure code) is favoured. Thus, for the MIMIC-III full label setting, we also report Micro-$F_1$ results on the diagnosis codes ($F_1$-diag) and procedure codes ($F_1$-proc) separately as in \cite{mullenbach-etal-2018-explainable}. 

Furthermore, we report the example-based metric, precision@$k$ as in \cite{mullenbach-etal-2018-explainable}, averaged over all the documents, where each precision score is the fraction of the true positive in the top-$k$ labels, having highest score $p_{dl}$, for the document $d$. The idea is to simulate the real-world scenario that the system recommending $k$ predicted medical codes and to evaluate the percentage of them being correct. The number of top-ranked labels $k$ were set as 8 for MIMIC-III, 5 for MIMIC-III-50 to be consistent to the study \cite{mullenbach-etal-2018-explainable}, and 1 for MIMIC-III-shielding, near to the average number of labels per document (see Table \ref{data_stat}).

\subsection{Main Results}
\label{main-results}
We report the mean and the standard deviation (i.e. the square root of variance) of the testing results of 10 runs with randomly initialised parameters for each model. The results of the MIMIC-III-50, MIMIC-III-shielding, and MIMIC-III datasets are shown in Table \ref{mimic-iii-50-results}, \ref{mimic-iii-shielding-results}, and \ref{mimic-iii-results}, respectively.

For the top 50 label dataset (MIMIC-III-50, see Table \ref{mimic-iii-50-results}), HLAN performed the best among all experimental settings, achieved significantly better Micro-AUC (91.9\%), Micro-$F_1$ (64.1\%), and Precision@5 (62.5\%) than the second best model, CNN. This shows the advantage of the hierarchical label-wise attention mechanisms for top-50 code prediction. With the same calibration threshold, the precision of HLAN is better than CNN absolutely by 15\% (73.2\% vs. 57.7\%), while recall is lower with a similar absolute value, indicating that tuning the threshold to balance precision and recall could further improve the $F_1$ scores.

\begin{table*}[th]
\caption{Results on MIMIC-III-50 dataset (50 labels)}\label{mimic-iii-50-results}
\footnotesize
\center
\begin{threeparttable}
\begin{tabular}{lllll|llll|l}
\cline{1-10}
                      & \multicolumn{4}{c}{Macro}                 & \multicolumn{4}{c}{Micro}                 & Top-$k$    \\
Model                 & AUC      & P        & R        & $F_1$       & AUC      & P        & R        & $F_1$       & P@5      \\
\cline{1-10}
CNN                   & 88.1$\pm$0.3 & 51.5$\pm$0.9 & \textbf{\underline{67.4$\pm$1.0}} & 58.4$\pm$0.5 & 90.9$\pm$0.2 & 55.6$\pm$1.1 & \textbf{\underline{71.2$\pm$0.9}} & 62.4$\pm$0.6 & 61.8$\pm$0.3 \\
+LE                   & \underline{88.3$\pm$0.3} & \underline{53.0$\pm$1.0}$^*$ & 66.7$\pm$1.5 & \textbf{\underline{59.1$\pm$0.5}}$^*$ & \underline{91.3$\pm$0.1}$^*$ & \underline{57.7$\pm$1.4}$^*$ & 70.4$\pm$1.4 & \underline{63.4$\pm$0.5}$^*$ & \underline{62.1$\pm$0.3} \\
\cline{1-10}
Bi-GRU                & 80.6$\pm$1.1 & 47.2$\pm$3.2 & 36.7$\pm$2.6 & 41.2$\pm$2.3 & 85.5$\pm$1.0 & \underline{58.1$\pm$3.2} & 45.8$\pm$2.2 & 51.2$\pm$1.9 & 51.3$\pm$1.7 \\
+LE                   & \underline{80.9$\pm$0.8} & \underline{47.3$\pm$2.0} & \underline{39.2$\pm$2.0}$^*$ & \underline{42.8$\pm$1.5} & \underline{85.8$\pm$0.7} & 57.5$\pm$2.2 & \underline{48.4$\pm$2.1}$^*$ & \underline{52.5$\pm$1.3}$^*$ & \underline{52.1$\pm$1.2} \\
\cline{1-10}
CNN+att               & 88.1$\pm$0.0 & 63.1$\pm$0.1 & \underline{48.4$\pm$0.2}$^*$ & \underline{54.8$\pm$0.2}$^*$ & 91.1$\pm$0.0 & 70.9$\pm$0.2 & \underline{53.1$\pm$0.2}$^*$ & \underline{60.7$\pm$0.1} & 60.8$\pm$0.1 \\
+LE                   & \underline{88.3$\pm$0.0}$^*$ & \underline{64.3$\pm$0.3}$^*$ & 46.0$\pm$0.1 & 53.6$\pm$0.1 & \underline{91.3$\pm$0.0}$^*$ & \underline{71.6$\pm$0.1}$^*$ & 52.5$\pm$0.1 & 60.6$\pm$0.1 & \underline{61.6$\pm$0.1}$^*$ \\
\cline{1-10}
HAN                   & 87.0$\pm$0.4 & \underline{61.7$\pm$2.7} & 46.3$\pm$2.3 & 52.8$\pm$1.1 & 90.1$\pm$0.3 & \underline{68.2$\pm$3.1} & 52.9$\pm$2.4 & 59.4$\pm$0.7 & 59.5$\pm$0.7 \\
+LE                   & \underline{87.3$\pm$0.4} & 61.3$\pm$3.2 & \underline{46.9$\pm$2.9} & \underline{53.0$\pm$1.1} & \underline{90.3$\pm$0.4} & 67.9$\pm$4.1 & \underline{54.2$\pm$2.8} & \underline{60.1$\pm$0.7} & \underline{59.9$\pm$0.8} \\
\cline{1-10}
HA-GRU                 & 85.3$\pm$1.3 & 59.3$\pm$2.5 & 43.1$\pm$4.0 & 49.9$\pm$3.4 & 89.2$\pm$0.9 & 69.5$\pm$0.6 & 48.7$\pm$4.1 & 57.2$\pm$2.8 & 57.9$\pm$1.7 \\
+LE                   & \underline{86.4$\pm$0.7}$^*$ & \underline{62.1$\pm$1.9}$^*$ & \underline{44.3$\pm$2.3} & \underline{51.7$\pm$1.9} & \underline{90.1$\pm$0.5}$^*$ & \underline{71.1$\pm$1.2}$^*$ & \underline{50.7$\pm$2.3} & \underline{59.1$\pm$1.4}$^*$ & \underline{59.5$\pm$1.0}$^*$ \\
\cline{1-10}
HLAN                  & 88.4$\pm$0.7 & 65.0$\pm$1.2 & \underline{51.0$\pm$2.6} & \underline{57.1$\pm$1.6} & \textit{91.9$\pm$0.4} & 72.9$\pm$0.8 & \underline{57.3$\pm$2.5} & \textit{\textbf{\underline{64.1$\pm$1.4}}} & \textit{\textbf{\underline{62.5$\pm$0.7}}} \\
+LE                   & \textbf{\underline{88.4$\pm$0.5}} & \textbf{\underline{65.5$\pm$1.5}} & 50.2$\pm$1.1 & 56.8$\pm$0.8 & \textit{\textbf{\underline{91.9$\pm$0.3}}} & \textbf{\underline{73.2$\pm$0.6}} & 56.9$\pm$1.0 & \textit{64.0$\pm$0.7} & \textit{62.4$\pm$0.6} \\
+sent split           & 86.9$\pm$0.5 & 63.6$\pm$1.3 & 47.8$\pm$2.4 & 54.5$\pm$1.7 & 90.4$\pm$0.3 & 71.5$\pm$1.2 & 53.8$\pm$2.1 & 61.4$\pm$1.2 & 60.2$\pm$0.7 \\
\cline{1-10}
\end{tabular}
\begin{tablenotes}
\item The results of better metric score between the model with label embedding initialisation (``+LE'') and the model \textit{not} using LE initialisation are \underline{underlined}, and the asterisk (*) further marks the paired two-tailed t-tests with .95 significant level ($p<0.05$) between them. The best result for each metric (column) is in \textbf{bold}. The AUC, $F_1$, and P@5 scores in HLAN models with \textit{italics} indicates their significantly improved results ($p<0.05$) over the second best model category (i.e. HLAN vs. CNN). The model with lower variance is preferred if the average scores are the same.
\end{tablenotes}
\end{threeparttable}
\end{table*}

For code related to high-risk patients for shielding during the COVID-19 pandemic (MIMIC-III-shielding, see Table \ref{mimic-iii-shielding-results}), results (of Micro-AUC) show that HLAN (96.9\%) and HAN (97.6\%) performed comparably to the best performed model, CNN (97.9\%). HLAN obtained a high value of precision@1, slightly below CNN by 1\% (81.2\% vs 82.2\%), while the difference was not significant ($p>0.05$). The better performance of CNN (or HAN) may be because that smaller datasets like MIMIC-III-shielding, with much fewer documents and labels (see Table \ref{data_stat}), tends to favour models with simpler architectures.

\begin{table*}[th]
\caption{Results on MIMIC-III-shielding dataset (20 labels)}\label{mimic-iii-shielding-results}
\center
\footnotesize
\begin{threeparttable}
\begin{tabular}{lllll|llll|l}
\cline{1-10}
                      & \multicolumn{4}{c}{Macro}                 & \multicolumn{4}{c}{Micro}                 & Top-$k$    \\
Model                 & AUC      & P        & R        & $F_1$       & AUC      & P        & R        & $F_1$       & P@1      \\
\cline{1-10}
CNN                   & \textbf{\underline{96.9$\pm$0.2}}$^*$ & 59.8$\pm$1.2 & 59.6$\pm$1.3 & 59.7$\pm$0.9 & \textit{\textbf{\underline{97.9$\pm$0.4}}}$^*$ & \underline{80.5$\pm$1.3} & 76.2$\pm$0.9 & \textit{\textbf{\underline{78.3$\pm$1.0}}}$^*$ & \textbf{\underline{82.2$\pm$0.8}} \\
+LE                   & 96.7$\pm$0.2 & \underline{60.4$\pm$2.8} & \textbf{\underline{60.4$\pm$2.3}} & \underline{60.4$\pm$2.3} & 97.6$\pm$0.3 & 78.8$\pm$2.5 & \textbf{\underline{76.4$\pm$1.8}} & \textit{77.5$\pm$0.7} & 81.6$\pm$0.9 \\
\cline{1-10}
Bi-GRU                & 91.9$\pm$1.4 & 57.4$\pm$3.1 & 43.4$\pm$2.2 & 49.4$\pm$2.2 & 93.6$\pm$0.8 & 77.9$\pm$2.8 & 58.5$\pm$1.9 & 66.8$\pm$1.2 & 72.2$\pm$1.6 \\
+LE                   & \underline{92.0$\pm$1.6} & \underline{58.6$\pm$1.5} & \underline{46.8$\pm$2.3}$^*$ & \underline{52.0$\pm$1.4}$^*$ & \underline{95.1$\pm$0.7}$^*$ & \underline{78.1$\pm$2.1} & \underline{61.8$\pm$2.7}$^*$ & \underline{68.9$\pm$1.6}$^*$ & \underline{75.1$\pm$2.0}$^*$ \\
\cline{1-10}
CNN+att               & 88.9$\pm$1.3 & 46.7$\pm$4.6 & 37.6$\pm$2.4 & 41.7$\pm$3.3 & 93.5$\pm$0.2 & \textbf{\underline{86.9$\pm$1.2}}$^*$ & 52.9$\pm$2.8 & 65.7$\pm$2.0 & 70.0$\pm$2.6 \\
+LE                   & \underline{95.5$\pm$0.0}$^*$ & \underline{62.1$\pm$2.2}$^*$ & \underline{48.4$\pm$1.9}$^*$ & \underline{54.4$\pm$2.0}$^*$ & \underline{96.1$\pm$0.0}$^*$ & 83.3$\pm$0.5 & \underline{61.4$\pm$0.6}$^*$ & \underline{70.7$\pm$0.3}$^*$ & \underline{77.7$\pm$0.3}$^*$ \\
\cline{1-10}
HAN                   & 96.0$\pm$1.4 & \textbf{\underline{66.4$\pm$2.7}} & \underline{58.2$\pm$2.0} & \textbf{\underline{62.0$\pm$2.0}} & 97.4$\pm$0.3 & 82.9$\pm$1.8 & \underline{68.7$\pm$2.4} & \underline{75.1$\pm$1.5} & 78.1$\pm$1.7 \\
+LE                   & \underline{96.4$\pm$1.3} & 65.2$\pm$2.1 & 56.5$\pm$2.9 & 60.5$\pm$2.3 & \underline{97.6$\pm$0.3} & \underline{83.4$\pm$1.2} & 68.2$\pm$2.0 & 75.0$\pm$1.2 & \underline{79.2$\pm$1.7} \\
\cline{1-10}
HA-GRU                 & 93.4$\pm$2.0 & \underline{60.9$\pm$3.9} & \underline{51.6$\pm$2.8} & \underline{55.8$\pm$3.1} & 96.7$\pm$0.4 & \underline{83.0$\pm$2.1} & 65.8$\pm$2.2 & \underline{73.4$\pm$1.6} & \underline{80.3$\pm$1.5} \\
+LE                   & \underline{93.9$\pm$2.0} & 59.2$\pm$4.3 & 49.7$\pm$4.2 & 54.0$\pm$4.1 & \underline{96.8$\pm$0.9} & 81.3$\pm$4.0 & \underline{66.3$\pm$4.1} & 73.0$\pm$3.6 & 79.1$\pm$4.3 \\
\cline{1-10}
HLAN                  & 93.5$\pm$2.5 & 59.8$\pm$2.9 & \underline{53.2$\pm$2.6} & \underline{56.3$\pm$2.4} & \underline{96.9$\pm$0.7} & 81.4$\pm$1.8 & \underline{69.0$\pm$2.9}$^*$ & \underline{74.6$\pm$1.6} & \underline{81.2$\pm$1.2} \\
+LE                   & 93.5$\pm$1.9 & 60.5$\pm$4.2 & 52.7$\pm$5.0 & 56.3$\pm$4.6 & 96.5$\pm$0.4 & \underline{81.8$\pm$2.8} & 65.6$\pm$4.0 & 72.7$\pm$3.1 & 79.8$\pm$3.0 \\
+sent split           & \underline{94.5$\pm$1.2} & \underline{60.9$\pm$2.1} & 51.7$\pm$3.1 & 55.8$\pm$2.3 & 96.3$\pm$0.2 & 81.4$\pm$2.2 & 64.4$\pm$2.5 & 71.9$\pm$1.7 & 77.8$\pm$2.6 \\
\cline{1-10}
\end{tabular}
\begin{tablenotes}
\item The results of better metric score between the model with label embedding initialisation (``+LE'') and the model \textit{not} using LE initialisation are \underline{underlined}, and the asterisk (*) further marks the paired two-tailed t-tests with .95 significant level between them. The best result for each metric (column) is in \textbf{bold}. The AUC, $F_1$, and P@1 scores in CNN models with \textit{italics} indicates their significantly improved results ($p<0.05$) over the second best model category (i.e. CNN vs. HAN or HLAN). The model with lower variance is preferred if the average scores are the same.
\end{tablenotes}
\end{threeparttable}
\end{table*}

In both MIMIC-III-50 and MIMIC-III-shielding, HA-GRU did not perform better than HLAN, this shows that the label-wise word-level attention mechanisms in HLAN further improved the performance. Also, surprisingly, the HLAN or HAN models with the real sentence split did not perform better (up to 2.8\% less Micro-$F_1$) than using text chunk ``sentences'' (of 25 continuous tokens) in all three datasets. This is probably because, with the sentence split setting, some tokens and sentences were lost during the padding procedure, which could significantly affect the performance.

For the full label setting (``MIMIC-III''), HAN has better results of Micro-AUC and precision@8 than the vanilla CNN and Bi-GRU, but worse than the CNN+att approach specifically tuned for this dataset. With label embedding initialisation, CNN+att+LE achieved significant best results on MIMIC-III (an Micro-AUC of 98.6\%). It is worth to further explore to enhance the scalability of HLAN so that it can process datasets with large label sizes. Also to note that results of the Macro-level metrics (averaging over labels) were dramatically lower than the Micro-level ones (calculated from document-label pairs), showing the strong imbalance of labels in MIMIC-III (see Section \ref{subsec:data}).

\begin{table*}[th]
\caption{Results on MIMIC-III dataset (8,922 labels)}\label{mimic-iii-results}
\center
\footnotesize
\begin{threeparttable}
\begin{tabular}{lp{1cm}p{1cm}p{1cm}p{1cm}|p{1cm}p{1cm}p{1cm}p{1cm}p{1cm}p{1cm}|p{1cm}}
\cline{1-12}
                     & \multicolumn{4}{c}{Macro}      & \multicolumn{6}{c}{Micro}                              & Top-$k$ \\
Model                & AUC      & P & R  & $F_1$      & AUC      & P & R   & $F_1$       & $F_1$-diag  & $F_1$-proc  & P@8   \\
\cline{1-12}
CNN                  & 81.8$\pm$0.7 & 4.5$\pm$0.4   & \underline{3.7$\pm$0.5} & 4.1$\pm$0.4 & 97.0$\pm$0.1   & 51.0$\pm$2.8    & \underline{36.9$\pm$1.7} & 42.8$\pm$0.9 & 41.1$\pm$1.0   & 50.7$\pm$0.9 & 59.6$\pm$0.5      \\
+LE               & \underline{82.4$\pm$0.4}$^{*}$ & \underline{4.7$\pm$0.4}   & 3.6$\pm$0.2 & \underline{4.1$\pm$0.2} & \underline{97.1$\pm$0.1} & \underline{53.0$\pm$2.6}    & 36.9$\pm$1.2 & \underline{43.4$\pm$0.6} & \underline{41.7$\pm$0.6} & \underline{51.3$\pm$0.9} & \underline{60.3$\pm$0.4}$^{*}$      \\
\cline{1-12}
Bi-GRU               & 83.5$\pm$1.6 & 4.9$\pm$0.2   & \underline{3.6$\pm$0.5} & 4.1$\pm$0.4 & 97.3$\pm$0.3 & 52.8$\pm$4.6  & 34.8$\pm$2.2 & 41.8$\pm$1.5 & 39.3$\pm$1.6 & 51.7$\pm$1.3 & 58.9$\pm$2.2      \\
+LE            & \underline{84.9$\pm$0.7}$^{*}$ & \underline{5.0$\pm$0.4}     & \underline{3.6$\pm$0.5} & \underline{4.2$\pm$0.5} & \underline{97.6$\pm$0.1}$^{*}$ & \underline{55.4$\pm$4.1}  & \underline{34.8$\pm$2.4} & \underline{42.6$\pm$1.5} & \underline{40$\pm$1.6}   & \underline{52.7$\pm$1.1} & \underline{60.3$\pm$1.8}      \\
\cline{1-12}
CNN+att              & 88.6$\pm$0.2 & 7.7$\pm$0.2   & 6.4$\pm$0.3 & 7.0$\pm$0.2   & 98.4$\pm$0.0   & \underline{62.8$\pm$0.3}$^{*}$  & 43.9$\pm$0.4 & 51.7$\pm$0.1 & 50.1$\pm$0.2 & 59.8$\pm$0.1 & 69.4$\pm$0.2      \\
+LE           & \textbf{\underline{90.2$\pm$0.0}}$^{*}$   & \textbf{\underline{9.3$\pm$0.1}}$^{*}$   & \textbf{\underline{8.0$\pm$0.1}}$^{*}$   & \textbf{\underline{8.6$\pm$0.1}}$^{*}$ & \textbf{\underline{98.6$\pm$0.0}}$^{*}$   & 61.8$\pm$0.4  & \textbf{\underline{45.6$\pm$0.1}}$^{*}$ & \textbf{\underline{52.5$\pm$0.1}}$^{*}$ & \textbf{\underline{50.7$\pm$0.1}}$^{*}$ & \textbf{\underline{60.7$\pm$0.1}}$^{*}$ & \textbf{\underline{69.7$\pm$0.1}}$^{*}$      \\
\cline{1-12}
HAN                  & \underline{88.5$\pm$0.1}$^{*}$ & \underline{5.4$\pm$0.2}$^{*}$   & \underline{2.7$\pm$0.2}$^{*}$ & \underline{3.6$\pm$0.2}$^{*}$ & 98.1$\pm$0.1 & 63.2$\pm$3.3  & \underline{30.0$\pm$1.1}$^{*}$   & \underline{40.7$\pm$0.7}$^{*}$ & \underline{37.0$\pm$0.7}$^{*}$   & \underline{52.6$\pm$0.9}$^{*}$ & \underline{61.4$\pm$1.3}$^{*}$      \\
+LE               & 88.2$\pm$0.2 & 5.1$\pm$0.2   & 2.4$\pm$0.1 & 3.3$\pm$0.2 & \underline{98.1$\pm$0.0}   & \textbf{\underline{63.2$\pm$1.0}}    & 27.6$\pm$1.1 & 38.4$\pm$1.0   & 34.8$\pm$1.2 & 50.6$\pm$1.0   & 59.6$\pm$0.6      \\
+sent split & 87.4$\pm$0.7 & 4.7$\pm$0.6   & 2.3$\pm$0.4 & 3.1$\pm$0.5 & 97.9$\pm$0.1   & 60.4$\pm$2.2  & 25.3$\pm$2.5 & 35.6$\pm$2.6 & 31.4$\pm$2.6 & 49.1$\pm$2.3 & 56.3$\pm$2.0 \\
\cline{1-12}
\end{tabular}
\begin{tablenotes}
\item The results of better metric score between the model with label embedding initialisation (``+LE'') and the model \textit{not} using LE initialisation are \underline{underlined}, and the asterisk (*) further marks the paired two-tailed t-tests with .95 significant level between them. The best result for each metric (column) is in \textbf{bold}. The model with lower variance is preferred if the average scores are the same.
\end{tablenotes}
\end{threeparttable}
\end{table*}

Injecting the code relations through label embedding consistently boosted the performance of automated medical coding. It is clear that most models were improved with label embedding initialisation (``+LE''). Models were affected to different extend by label embedding: CNN+att model was mostly improved with ``+LE'' (an increase of 6.6\% Macro-AUC on MIMIC-III-shielding), the rest models (CNN, Bi-GRU, HA-GRU) being relatively less affected, while there was no significant improvement for HLAN or HAN on the datasets. This may due to the fact that the prior layers, e.g. hierarchical layers and the label-wise attention layers, could already learn some of the label relations. We thus further analyse the LE-initialised layers in Section \ref{le_analysis} to understand the effect of label embedding initialisation. Besides, most metrics with the ``+LE'' models also have higher stability (i.e. reduced variance); and low variance is an essential characteristic to deploy a model in the clinical setting.

\subsection{Result for each label}
\begin{figure*}[t]
  \centering
  \includegraphics[width=0.48\textwidth]{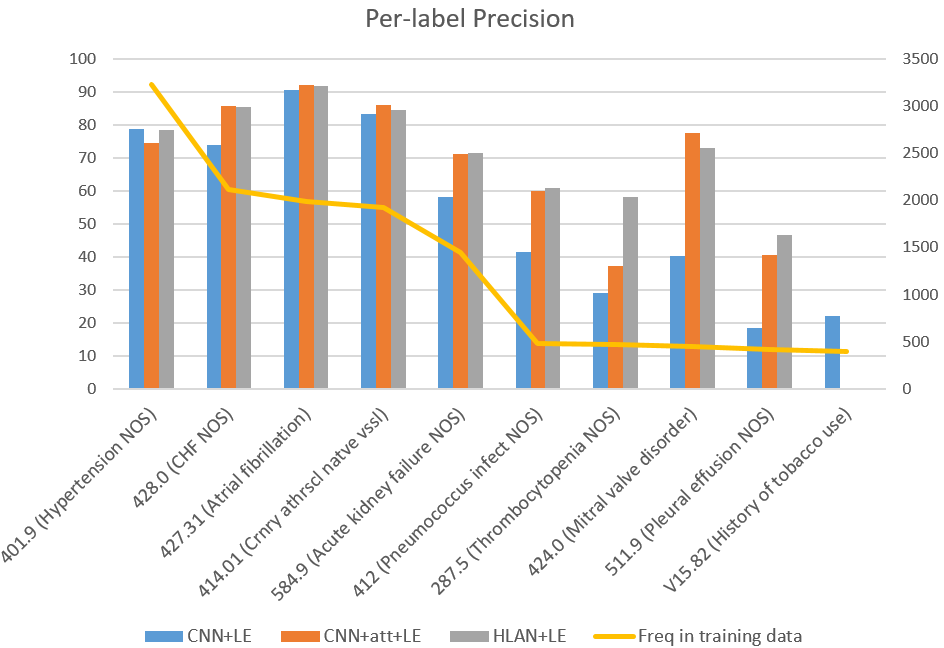}
  \includegraphics[width=0.48\textwidth]{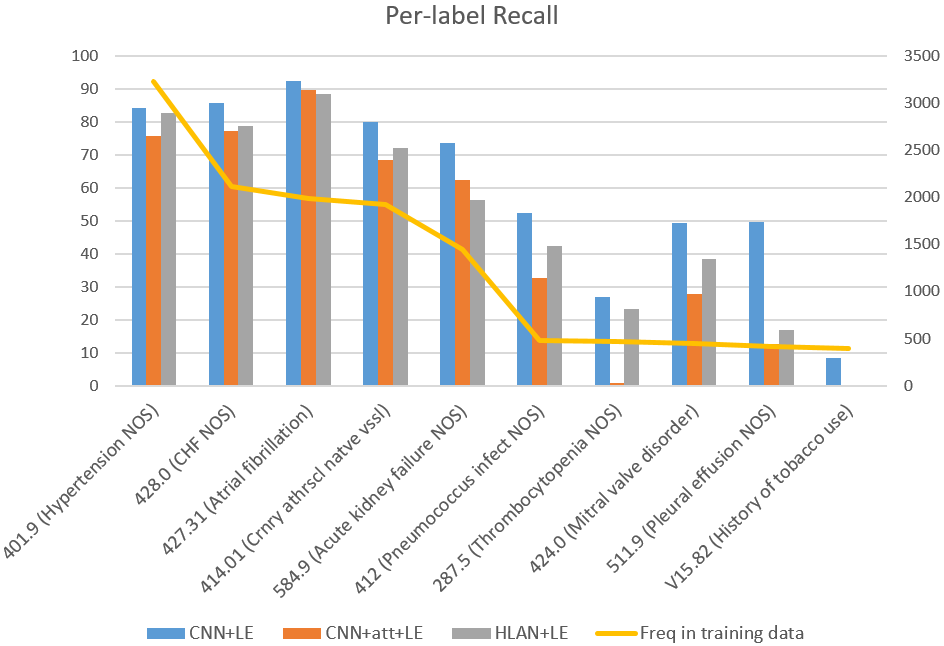}
  \caption{Precision and recall of the five most and the five least frequent ICD-9 diagnosis codes in the MIMIC-III-50 dataset. The bar chart (with the left y-axis) shows the metric score, while the line chart (with the right y-axis) shows the number of occurrences or the frequency (``Freq'') of the label in the training data.}\label{per-label-results}
\end{figure*}

Apart from the overall performance of the models, it is also essential to see how the models perform regarding each medical code. Figures \ref{per-label-results} show the precision and recall of the five diagnosis codes having the highest and the lowest frequencies in MIMIC-III-50. For this analysis, we selected the three best performing models, CNN, CNN+att, and HLAN, all with label embedding initialisation (``+LE''), in terms of AUC metrics for MIMIC-III-50 (see Table \ref{mimic-iii-50-results}). We provide the full per-label results of HLAN+LE with the MIMIC-III-50 and MIMIC-III-shielding datasets in Table S1-S2 in the supplementary material.

In Figures \ref{per-label-results}, we can observe that the overall trend of performance is generally consistent to, while not solely dependent on, the label frequency in the training data. For the five most frequent labels, the models achieved around 70\%-90\% precision and recall. For example, in terms of precision, HLAN obtained highest to 91.7\% for 427.31 (Atrial fibrillation) and lowest to 71.4\% for 584.9 (Acute kidney failure). For the five least frequent labels, the results were much worse due to the fewer training data for the labels and the imbalance issue. For precision, HLAN generally performs better than CNN and CNN+att, especially there is a significant gap for low frequent labels; while for recall, CNN outperforms the other two models. We also note that the precision and recall could be tuned in favour of only one of them through changing the calibration threshold $\mathit{Th}$ (now set as the default value, 0.5), considering the need and the preference of the coding work when deploying the model to support coding professionals.

\subsection{Model Explanation with Hierarchical Label-wise Attention Visualisation}
\label{sec:anal-att-viz}

\begin{figure*}[t]
  \includegraphics[width=0.95\textwidth]{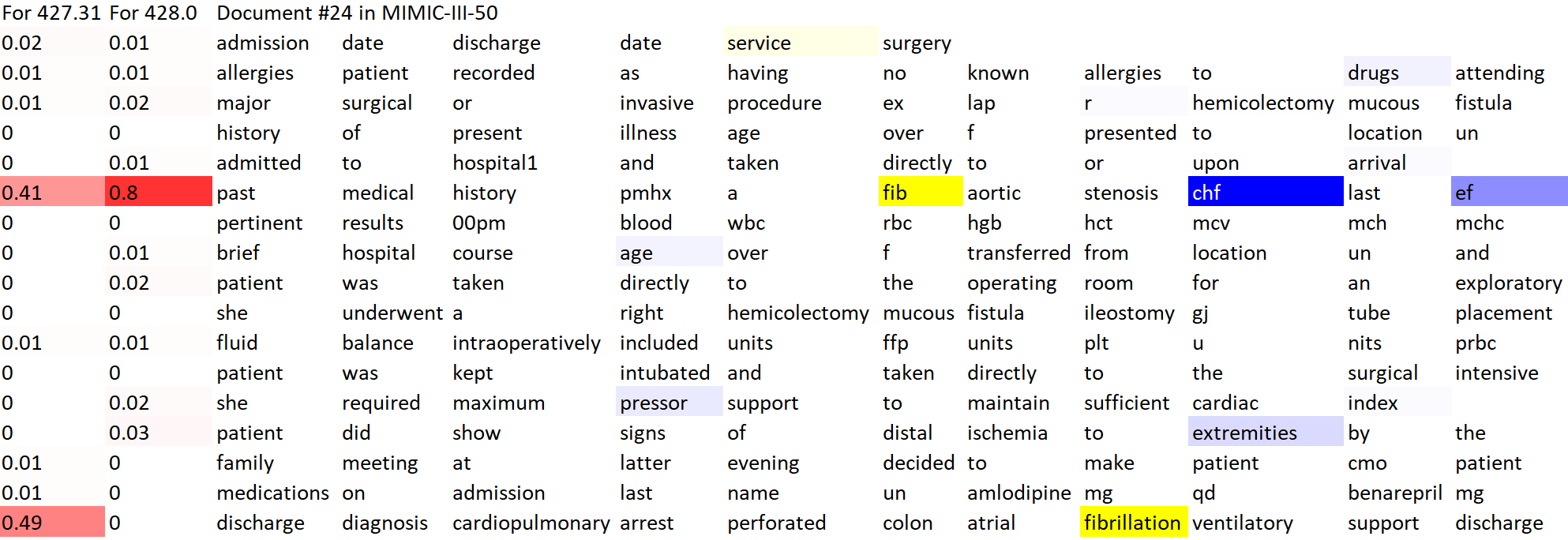}
  \caption{An example of interpretation using attention visualisation from the Hierarchical Label-wise Attention Network (HLAN), the chosen example is a random document (index 24) in the MIMIC-III-50 dataset with two true positive labels, ICD-9 code 427.31 (Atrial fibrillation) and 428.0 (Congestive heart failure, unspecified). The two \colorbox{red}{red} columns show the sentence-level attention scores for the two codes respectively. The tokens highlighted by \colorbox{yellow}{yellow} (for code 427.31) or \colorbox{blue}{\textcolor{white}{blue}} (for code 428.0) show the importance of them based on the value of sentence-weighted word-level attention scores. The deeper the colour, the higher the (sentence-weighted) attention scores, and thus the more important the highlight words or sentences contributes to the model prediction. Only the first part (11 tokens) of each sentence was shown for a clearer display.}\label{hlan_att_viz}
\end{figure*}

A critical requirement of the clinical use of automated medical coding systems is their explainability or interpretability. We propose to use label-wise word-level and sentence-level attention mechanisms in HLAN to enhance the explainability of the model. The learned word-level and sentence-level attention scores for the label $y_l$ are $\alpha_{wl} \in (0,1)$ and $\alpha_{sl} \in (0,1)$ (see Equations \ref{word-att} and \ref{sent-att}, respectively). For a more concise visualisation, we propose a \textit{sentence-weighted} word-level attention score $\tilde{\alpha}_{wl}$ to only highlight the words from the salient sentences. This adapted word-level attention score is calculated as $\tilde{\alpha}_{wl}$ = $\mu \alpha_{\tilde{s}l}\alpha_{wl}$, where $\alpha_{\tilde{s}l}$ is the attention score of the sentence where the word is belong to and $\mu$ is a hyperparameter to control the magnitude of the final weighted attention score. A greater $\mu$ will result in highlighting more words in the clinical note and we empirically set $\mu$ as 5. We clip the value of $\tilde{\alpha}_{wl}$ to 1 if it is above 1.

An example attention visualisation for a random document (number 24) in MIMIC-III-50 using the model HLAN+LE with the parsed sentences (``+sent split'') is shown in Figure \ref{hlan_att_viz}. The two columns on the left visualise the sentence-level attention scores $\alpha_{sl}$ for the two codes, 427.31 (Atrial fibrillation) and 428.0 (Congestive heart failure, unspecified), respectively. The highlighted sentences are corresponding to the sections ``past medical history'' and ``discharge diagnosis'' in the discharge summary. This is in line with our intuition that the key diagnosis information is likely to be contained in the two sections. The words are highlighted according to the adapted word-level attention score $\tilde{\alpha}_{wl}$. Words related to the code 427.31 is highlighted in yellow and for 428.0 in blue. It is clear that the most salient words are highlighted, and the model successfully recognised the abbreviations and alternative short forms commonly used by clinicians in the clinical note, for example ``a fib'' as a short form of atrial fibrillation and ``chf'' as the abbreviation of Congestive Heart Failure. This shows that the proposed HLAN model can learn to recognise the strongly correlated words (e.g. ``a fib'') related to the label (e.g. the code 427.31) with label-wise attention mechanisms, even given the fact that the label description (i.e. the knowledge that 427.31 is ``Atrial Fibrillation'') were not fed into the model during training. Other relevant words are highlighted, e.g., ``ef'' (short for Ejective Fraction), ``pressor'', and ``extremities'', which show a correlation to the code 428.0 while not indicating a causal relation to the diagnosis. We also note that the highlighted words like ``age'' and ``drugs'' were too general, which could not be directly related to the diagnosis from a clinician's point of view. This may be related to the peaky distribution of the softmax (normalised exponential) function to form the attention scores (see Equation \ref{word-att}), paying the most of the attention to only a few (one or two) words in a long sentence.

\subsection{Comparison of Model Explanations}
Following the previous section, we further qualitatively analyse and compare the interpretability of the HLAN model and other baseline models. Table \ref{tp-interpretability-results} shows how CNN, CNN+att, HAN, HA-GRU, and HLAN, all with label embedding initialisation, highlight the ``important'' part of a random document (number 24) to predict two different labels (427.31 and 428.0). The CNN\footnote{We further normalised the scores of $n$-grams in CNN based on max-pooling from \cite{mullenbach-etal-2018-explainable} to probabilities, to be comparable to the attention scores in other models.} and CNN+att chose the most salient $n$-grams based on the max-pooling and the attention mechanism, respectively \cite{mullenbach-etal-2018-explainable}. HLAN and its downgraded models, HA-GRU and HAN, alternatively, highlighted the important sentences and words. The distinction is that HAN has the same highlights of the same document for different labels (columns in Table \ref{tp-interpretability-results}), and HA-GRU has the same word-level but different sentence-level highlights across labels, while HLAN can highlight the most salient words and sentences for different labels. This gives HLAN the most comprehensive interpretability among the models.

\begin{table*}[th]
\caption{Comparison of model interpretability across deep learning models of true positive predictions on a random document (index 24) in the MIMIC-III-50 dataset}\label{tp-interpretability-results}
\footnotesize
\begin{threeparttable}
\begin{tabular}{lp{7.5cm}||p{7.5cm}}
\cline{1-3}
Model                    & doc-24 to predict 427.31 (\textbf{Atrial fibrillation})                                                                                                                                                                                                                                                                                                                                                & doc-24 to predict 428.0 (\textbf{Congestive heart failure, unspecified})                                                                                                                                                                                                                                                             \\
\cline{1-3}
\multirow{3}{*}{CNN+LE}  & $n$-gram-1 (0.105): admission date discharge \textbf{date service surgery allergies patient} recorded as having no known...                                                                                                                                                                                                                                                                                                                          & $n$-gram-1 (0.096): ...surgical intensive care unit she \textbf{required maximum pressor support} to maintain sufficient cardiac index...\\
\cline{2-3}
                         & $n$-gram-2 (0.083): ...surgical or invasive procedure ex lap r \textbf{hemicolectomy mucous fistula ileostomy gj tube} placement history of present illness...                                                                                                                                                                                                                                                                                                                                                          & $n$-gram-2 (0.075): ...past medical history pmhx a \textbf{fib aortic stenosis chf} last ef in osteoporosis reflux...\\
                         \cline{2-3}
                         & $n$-gram-3 (0.075): ...presented to location un with \textbf{perforated viscous hd stable} upon transfer to location un...
                         & $n$-gram-3 (0.071):...on ventilation support family meeting \textbf{at latter evening decided} to make patient cmo patient...\\
\cline{1-3}
\multirow{2}{*}{CNN+att+LE} & $n$-gram-1 (0.026): ...upon arrival past medical history \textbf{pmhx a fib aortic stenosis chf last ef in osteoporosis reflux doctor first name hx appendectomy many} years ago social history non...                                                                                                                                                                                                                                                                             & $n$-gram-1 (0.017): ...pressor support to maintain sufficient \textbf{cardiac index patient did show signs of distal ischemia to extremities by} the afternoon urine output post...                                                                                                                                                                                                                                          \\
\cline{2-3}
                         & $n$-gram-2 (0.023): ...diagnosis cardiopulmonary arrest perforated colon \textbf{atrial fibrillation ventilatory support discharge condition death discharge instructions none followup} instructions none                                                                                                                                                                                                                                                                           & $n$-gram-2 (0.011): ...past medical history pmhx a \textbf{fib aortic stenosis chf last ef in osteoporosis reflux doctor first name hx appendectomy many years} ago social history non contributory...                                                                                                                                                                                                              \\
\cline{1-3}\cline{1-3}
\multirow{3}{*}{HAN+LE}  & \multicolumn{2}{p{15cm}}{sent-1 (0.34): medical history pmhx a \textbf{fib(0.374)} aortic stenosis \textbf{chf(0.206)} last ef in osteoporosis \textbf{reflux(0.097)} doctor first name hx appendectomy \textbf{many(0.28)} years ago social history non contributory}\\
\cline{2-3}
                         & \multicolumn{2}{p{15cm}}{sent-2 (0.18): arthritis fosamax q \textbf{week(0.039)} \textbf{coumadin(0.282)} qd discharge medications none discharge disposition expired discharg diagnosis \textbf{cardiopulmonary(0.019)   arrest(0.109) perforated(0.134) colon(0.119) atrial(0.173) fibrillation(0.023) ventilatory(0.047)} support discharge condition death}\\
\cline{2-3}
                         & \multicolumn{2}{p{15cm}}{sent-3 (0.11): \textbf{admission(0.201) date(0.263) discharge(0.05) date(0.055) service(0.075) surgery(0.118) allergies(0.062) patient(0.013) recorded(0.054)} as having no known allergies to drugs attending first \textbf{name3(0.021) lf(0.02)} chief complaint perforated \textbf{bowel(0.046)} major}\\
\cline{1-3}
HA-GRU+LE                & sent-1 (0.62): arthritis fosamax q week \textbf{coumadin(0.06)} qd   discharge medications none discharge disposition   expired discharge diagnosis cardiopulmonary arrest   perforated colon atrial \textbf{fibrillation(0.94)} ventilatory   support discharge condition death & Did not predict 428.0 (i.e. false negative)                                                                                                                                                                                                                                                                                                   \\
\cline{1-3}
\multirow{2}{*}{HLAN+LE} & sent-1 (0.54): arthritis fosamax q week coumadin qd   discharge medications none discharge disposition   expired discharge diagnosis cardiopulmonary arrest   perforated colon \textbf{atrial(1.0)} fibrillation ventilatory   support discharge condition death   & sent-1 (0.71): medical history   pmhx a fib aortic stenosis \textbf{chf(1.0)} last   ef in osteoporosis reflux doctor first   name hx appendectomy many years ago social   history non contributory \\
\cline{2-2}
                         & sent-2 (0.18): medical   history pmhx a \textbf{fib(1.0)} aortic stenosis chf   last ef in osteoporosis reflux doctor   first name hx appendectomy many years ago   social history non contributory &\\
\cline{2-3}
\multirow{2}{*}{+sent split} & sent-1 (0.49): discharge diagnosis cardiopulmonary arrest perforated colon atrial \textbf{fibrillation(1.0)} ventilatory support discharge condition death discharge instructions none followup instructions none   & sent-1 (0.8):  past medical history pmhx a fib aortic stenosis \textbf{chf(0.888)} last \textbf{ef(0.112)} in osteoporosis reflux doctor first name hx appendectomy many years ago social history non \\
\cline{2-2}
                         & sent-2 (0.41): past medical history pmhx a \textbf{fib(1.0)} aortic stenosis chf last ef in osteoporosis reflux doctor first name hx appendectomy many years ago social history non &\\
\cline{1-3}
\end{tabular}
\begin{tablenotes}
\item$^{*}$ CNN and CNN+att suggested top $n$-grams, while HAN, HA-GRU, and HLAN suggested key sentences (``sent-'') and words in the sentences. ``+sent split'' denotes the HLAN model using real sentence splits. The numbers in the parentheses are the attention scores (e.g. for HLAN, $\alpha_w$ and $\alpha_s$) in the models.
\item$^{**}$ For CNN and CNN+att, some of the suggested top-3 $n$-grams were combined together if any of them overlapped; up to five tokens before and after the top $n$-grams were also displayed.
\item$^{***}$ For HAN, HA-GRU, and HLAN, the sentences were selected by those with sentence-level attention scores above 0.1 and the words were selected by those with the word-level attention scores above 0.01. Both HAN and HA-GRU predicted 427.31 but not 428.0. The word- and sentence-level attention weights of HAN are shared for all labels, therefore the interpretation is the same for both columns.
\end{tablenotes}
\end{threeparttable}
\end{table*}

Compared to CNN, we observe that CNN+att generated a more relevant set of $n$-grams. This is in accordance with the conclusion in \cite{mullenbach-etal-2018-explainable}. We also found that the attention weights from CNN are unstable, i.e. the suggested $n$-grams from CNN were not the same among different runs. Compared to the interpretation with $n$-grams, highlighting the key sentences and words can produce a more comprehensive interpretation, as the latter is based on the whole hierarchical structure of a document. Especially with the sentence parsing (``HLAN+LE+sent split'', see the last row in Figure \ref{tp-interpretability-results}, corresponding to the visualisation in Figure \ref{hlan_att_viz}), we can clearly see which sections of the discharge summary, along with words, contribute more to predict the label.

\begin{table*}[th]
\caption{Examples of false positives of HLAN (with label embedding ``+LE'' and sentence splitting ``+sent split'') on MIMIC-III-50 and MIMIC-III-shielding}\label{fp-interpretability-results}
\footnotesize
\begin{tabular}{lp{2.7cm}p{7cm}p{2.7cm}}
\cline{1-4}
Document index (dataset)                          & \textit{False positive} ICD-9 code    & Explanation with the most relevant sentences and words by attention scores                                                                                                                                                                                                                      & potential reason                                                     \\
\cline{1-4}
\multirow{3}{*}{doc-68   (MIMIC-III-50)}     & \multirow{3}{3cm}{427.31 (Atrial fibrillation)} & sent-1 (0.32): discharge diagnosis septic shock due to ascending   cholangitis choledocholithiasis atrial \textbf{fibrillation(1.0)} with rapid   ventricular response pulmonary emboli deep venous thrombosis upper gi bleed   peptic ulcer & \multirow{3}{3cm}{missed coding}                       \\
\cline{3-3}
                                             &                         & sent-2 (0.31): past medical history   recent pe dvt \textbf{afib(1.0)} htn hypotension hypothyroidism cad mild chf                                                                                                                           &                                                                      \\
\cline{3-3}
                                             &                         & sent-3 (0.25): she was found to have   bilateral pe s and new \textbf{afib(1.0)} and started on coumadin                                                                                                                                     &                                                                      \\
\cline{1-4}
doc-19 (MIMIC-III-50)                        & 401.9 (Hypertension NOS, or Unspecified essential hypertension)                   & sent-1 (0.84): decision made to proceed with primary right total knee   arthroplasty past medical history \textbf{htn(1.0)} asthma allergies diabetes social   history nc family history nc                                                  & \multirow{3}{3cm}{past disease or missed coding}      \\
\cline{1-4}
\multirow{2}{*}{doc-1 (MIMIC-III-shielding)} & \multirow{2}{3cm}{416.0 (Primary pulmonary hypertension)}  & sent-1 (0.45): brief hospital course year old female with h o mild   alzheimer s disease cea in \textbf{htn(0.177) elev(0.145) lipids(0.659)}   bladder ca who presents as a transfer                                   & \multirow{3}{3cm}{subtle difference in language (regarding the type of hypertension)}                                                                     \\
\cline{3-3}
                                             &                         & sent-2 (0.36): past medical history mild   alzheimer s disease l cea in \textbf{htn(0.284) elev(0.167) lipids(0.518)}   bladder ca no known metastasis                                                                          &                                                                      \\
\cline{1-4}
\multirow{3}{*}{doc-65 (MIMIC-III-shielding)} & \multirow{3}{3cm}{197.0 (Secondary malignant neoplasm of lung)}  & sent-1 (0.31): brief hospital course yo man with history of  \textbf{metastatic(1.0)} pancreatic cancer was admitted with dyspnea new ascites and   profound hyponatremia                                                   & \multirow{3}{3cm}{subtle difference in language (regarding the type of secondary cancer), imbalance of vocabularies or diseases in the training data}      \\
\cline{3-3}
                                             &                         & sent-2 (0.3): history of present illness   yo cantonese and spanish speaking male with \textbf{metastatic(1.0)} pancreatic cancer   was admitted from the ed with dyspnea altered mental status and                                          &                                                                      \\
                                             \cline{3-3}
                                             &                         & sent-3 (0.1): \textbf{metastatic(1.0)} pancreatic   cancer evidence of progression of ct abdomen pelvis                                                                                                                                      &                                                                      \\
\cline{1-4}
\multirow{2}{*}{doc-95 (MIMIC-III-shielding)} & \multirow{2}{3cm}{280.00 (Neutropenia, unspecified)} & sent-1 (0.24): she has since been found to have a rising ldh and   hypercalcemia and decided with her \textbf{local(0.538)} oncologist dr   first \textbf{name8(0.448)} namepattern2 name stitle to                       & \multirow{2}{3cm}{subtle difference between the predicted label 280.00 and the ground truth label 280.03 (Drug induced neutropenia)} \\
                                             &                         & sent-2 (0.17):  at presentation on she developed   hypercalcemic with a \textbf{calcium(1.0)} of an elevated ldh                                                                                                                             & \\
\cline{1-4}
\end{tabular}
\end{table*}

It is also interesting to see how the proposed model interpret when it predicted a medical code not previously assigned by the coding professionals. We selected some representative ``false positive'' results from the HLAN+LE model with sentence splitting in Table \ref{fp-interpretability-results}. We presented the prediction results and the highlighted explanations to an experienced clinician to validate and deduce the potential reason for the error. In Table \ref{fp-interpretability-results}, we observe that the model can explain the predictions with key sentences and words, therefore it is easier for us to know where there may have been a problem. For example, for the first two rows, ``doc-68'' and ``doc-19'' in MIMIC-III-50, the highlighted words and sentences are quite relevant to the non-coded, ``false positive'' ICD-9 code, indicating that there might have been missed coding or the disease was a past disease of the patient.

The false positives in ``doc-1'' and ``doc-65'' in MIMIC-III-shielding are errors related to the wrong correlations learned from the data, particularly regarding the high granularity and subtle difference among sub-type diseases. In ``doc-1'', the highlighted words ``htn elev lipids'' show that the patient has a certain type of hypertension, but does not necessarily mean the predicted code 416.0 for ``Primary pulmonary hypertension''. In ``doc-65'', the strongly highlighted word ``metastasis'' actually is related to pancreatic cancer, rather than the more common lung cancer as predicted. This wrong correlation may be due to the imbalance of vocabularies in the training data: there are 94 (about 2\% out of 4,574) discharge summaries in the training data where ``pancreatic'' and ``metasta'' appeared together in MIMIC-III-shielding, while there are significant more discharge summaries (661, about 14\%) where ``lung'' and ``metasta'' appeared together.

The error in the last example was due to the subtle difference between two codes (280.00 vs 280.03). We noticed some unexpected highlights (e.g. the local oncologist's name code) in the last example (``doc-95'' in MIMIC-III-shielding). This may be related to that ``hypercalcemia'' (appeared in both sent-1 and sent-2) can be caused by cancer, while neutropenia can be caused by treatments like cancer chemotherapy. The word ``chemotherapy'' was highlighted in another sentence with an attention weight 0.09 (not presented as below the 0.1 threshold) and the word ``neutropenia'' in the document was not included during the padding process. While it is very likely that the neutropenia was induced by the drug for chemotherapy (that is, 280.03, Drug induced neutropenia), we did not find direct words in the report to point the cause of the disease (thus 280.00, Neutropenia, unspecified, is also appropriate).

In general, we observe that the label-wise attention mechanisms in the HLAN model can provide a more comprehensive explanation to support the predictions. For wrong or non-coded predictions, the explanations through highlighted sentences and words can help us better understand the problem. This provides an essential reference to help coding professionals use the system and help engineers fix the problems for the next system iteration.

\subsection{Analysis of Label Embedding Initialisation}
\label{le_analysis}
We previously visualised the label embedding from the MIMIC-III dataset reduced to two dimensions using T-SNE, in Figure \ref{LE-plot}. The visualisation intuitively shows how the label embedding can capture the correlations among ICD-9 codes derived from the coding practice in the clinical setting.

It is also interesting to know, after the dynamic update during training, how the weights in the initialised layers (the final projection layer and the attention layer) preserve the semantics of the label embedding, and why, in a few cases, LE did not result in a significant improvement. We thus extracted the weights in the learned layers and measured their similarity to the original label embedding. Based on the idea of label similarity, we calculated the top-10 similar labels for every label based on the pairwise cosine similarity of the rows in the initialised layer weights (e.g. rows such as $w_j$, $w_k$ in $W$ in Equation \ref{projection} or rows $V_{wl}$ in $V_w$ in Equations \ref{word-att}), and also the top-10 similar labels from original label embedding $E$, and then to see to what extent the two sets of ``top-10 similar labels'' overlap. We used the Jaccard Index to measure the degree of overlap between the two sets for each label, which is the size of the intersection divided by the size of the union of the two sets. We averaged the Jaccard Index over the labels. Thus the final metric reflects how the layers can retain the semantics, i.e. label similarities, of the label embedding $E$. We also used the models without the LE initialisation as the control group and calculated this averaged Jaccard Index from their layers for comparison.

\begin{figure*}[h]
  \centering
  \includegraphics[width=0.7\textwidth]{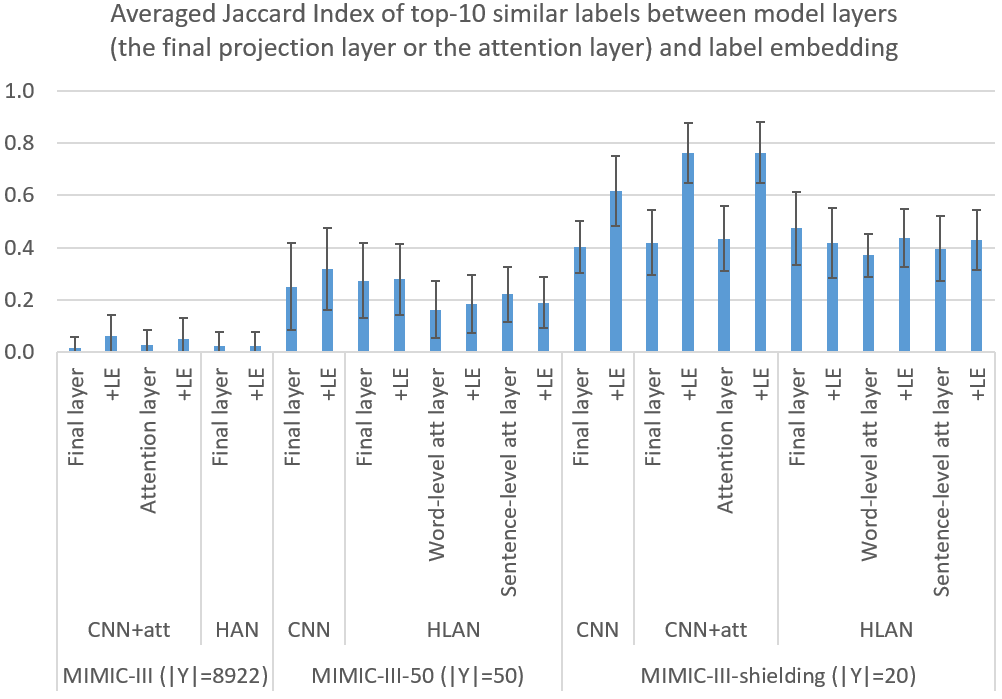}
  \caption{Averaged Jaccard Index between the sets of top-10 similar labels derived from the layers (final projection layer and label-wise attention layers with or without label embedding initialisation) and from the label embedding (LE). The higher the averaged Jaccard Index, the more similar the overall semantics between the layer weights and the pre-trained label embedding. Error bars show the standard deviation over the labels. Representative models are selected for all the three datasets. ``+LE'' means label embedding initialised for the layer indicated in the closest left bar.
  }\label{LE-init-analysis}
\end{figure*}

The results are displayed in Figure \ref{LE-init-analysis}. We selected several representative models that either were significantly improved with LE initialisation approach or did not improve with LE according to the results in Tables \ref{mimic-iii-50-results}-\ref{mimic-iii-results}. The experiment shows that weights in the final projection layer (and the label-wise attention layer, if applied) with LE initialisation (``+LE'') can capture further label similarities from the label embedding. We also observed a strong correlation between the performance improvement with LE (see Tables \ref{mimic-iii-50-results}-\ref{mimic-iii-results}) and the increase of averaged Jaccard Index with LE initialisation (i.e. the extent that the initialised layers captures the semantics of LE after training, as reflected in Figure \ref{LE-init-analysis}). The models which are more enhanced by LE (for example, CNN+att, with 6.6\% improvement of Macro-AUC with ``+LE'' in Table \ref{mimic-iii-shielding-results}) have a greater averaged Jaccard Index compared to the models without LE (0.76 vs. 0.42-0.43) in Figure \ref{LE-init-analysis}. On the contrary, the models which were \textit{not} improved with LE, e.g. HLAN and HAN, for automated coding, also, was also \textit{not} affected by LE in terms of the averaged Jaccard Index. The less effect of LE on HLAN and HAN may be because the hierarchical attention layers (especially with the label-wise attention mechanisms) could already model certain label correlations through the document-level matching process. In overall, the analysis supports the idea that LE initialisation, capturing the label correlations, is a key factor to enhance automated coding with deep learning based multi-label classification. Since LE can be visualised after dimensionality reduction (see Figure \ref{LE-plot}), this further serves as a mean to help explain the overall model.

\section{Discussion}
\label{discussion}
We have presented the results on the three datasets and analysed the interpretability of models and the layers initialised with label embeddings.

The main advantage of HLAN lies in its model explainability, based on the label-wise word-level and sentence-level attention mechanisms. The qualitative comparison of model explanation suggests that the highlighted key words and sentences from HLAN tend to be more comprehensive and more accurate than, those from HAN or HA-GRU and the $n$-gram explanation from the CNN related models. Such explainable highlights can be particularly helpful when medical coding professionals need to locate the essential part of a long clinical note. When the model suggests a code, its accompanying explanation could be served as a reference for professionals to validate whether the code should be included. This has the potential to build the users' trust in the deep learning model and help identify missed and erroneous coding.

The label embedding initialisation approach boosted the performance and reduced the variance of most models. The method is efficient, not requiring further model parameters. It is independent of the neural encoders, and can thus be applied to various deep learning models for multi-label classification. Our analysis on the LE initialised layers show that they can preserve the semantics in the pre-trained label embeddings and therefore can better capture the label similarity from the data. This further contributes to the explainability of the overall approach. There are a few exceptions that LE did not improve the performance, this may be due to the fact that the hierarchical layers can already model certain label correlations when optimising the document-label matching.

In terms of the performance, for MIMIC-III-50, the HLAN model with LE achieved significant better micro-level AUC (91.9\%) and $F_1$ score (64.1\%) than the previous state-of-the-art models; for MIMIC-III-shielding, HLAN and HAN performed comparably to CNN (all around 97-98\% micro-level AUC); for MIMIC-III, the previous state-of-the-art model CNN+att was significantly boosted by LE initialisation, achieving best AUC and $F_1$ scores (Micro-level AUC of 98.6\% and Micro-level $F_1$ of 52.5\%).

It is worth nothing that the higher comprehensiveness in explanation from HLAN is at the cost of further memory requirements and the training time\footnote{The estimated training and testing time of the models are in Table S3 in the Supplementary Material.}. Thus, in practice, if there are only limited computational resources (e.g. a single GPU with 12GB memory), we suggest training HLAN with a fewer number of codes, e.g. equal or less than 50, in a sub-disease domain or for specific tasks (i.e. shielding-related diseases during COVID-19) that require higher model explainability for decision making. We also notice that the vanilla CNN can be trained relatively faster with significantly less memory requirement; HAN and HA-GRU can also be applied as ``downgraded'' alternatives of HLAN for tasks with larger label sizes. It is also worth to explore to optimise the implementation and to distil the model of HLAN to enable its application to large label sizes.

While training deep learning models can be slow, during the testing phase, the trained models perform reasonably efficient for real-time inference. On average, it requires less than 1/3 second (330 milliseconds) to assign ICD codes with explainable highlights for a discharge summary with a CPU server using HLAN trained from MIMIC-III-50; and the CNN related models can process even faster (see Table S3 in the Supplementary Material). This allows the efficient use of the models in real-time for automated coding.

Also, the calibration threshold (default as 0.5) could be tuned to adjust the precision and recall of the system when deploying it to a coding department. While high precision is obtained when suggesting a few top-ranked predictions, a system with a higher recall can help the coding professionals to prevent missed coding. A higher recall can be achieved by using a lower calibration threshold, e.g. 0.3-0.4. The results are also highly varied across labels, as seen in the per-label results in Figures \ref{per-label-results}. A domain for future studies is therefore to investigate few-shot or zero-shot learning for the rare labels and we noticed one recent related work in \cite{rios-kavuluru-2018-shot}, which is based on ICD-9 hierarchies and descriptions to better predict rare labels.

The results demonstrate the usefulness of label embedding to boost coding performance for most models. In this work, the label embedding was trained with the label sets in the training data using the Continuous Bag of Words algorithm in word2vec. Thus, it encodes the similarity of medical codes derived from the real-world coding practice in the critical care unit of the US hospital. This knowledge is distinct from the ICD-9 hierarchy, as visualised in Figure \ref{LE-plot} and there may be contradictions between them. The advantage of the former is that it directly learned the label correlation from the existing hospital and does not require external knowledge. There are recently more studies on leveraging the hierarchy for medical coding as in \cite{falis-etal-2019-ontological,cao-etal-2020-hypercore}. One future direction is thus to combine the local knowledge with external knowledge for the task.

The analysis of false positives in the model (see Section \ref{sec:anal-att-viz} and Table \ref{fp-interpretability-results}) suggests further research in the area of automated medical coding. The errors are likely due to missed coding, past medical history rather than present diseases, nuances of language variations, imbalanced vocabularies, and high label granularity. The highlighted sentences and words helped us better determine the cause of the problems. Since missed coding is very common in real-world practice, as also pointed out recently in \cite{searle2020}, it is worth to adapt the current algorithms to capture missing labels and emerging new labels. Information on the report template may further help the model select the relevant part of a discharge summary and differentiate a present disease from a past disease. Unable to capture the subtle variations or labels is potentially related to wrong correlations learned from the imbalance of vocabularies and labels in the dataset. This may be addressed by incorporating various external knowledge.

\section{Conclusion}
\label{conclusion}
In this paper, we examined the existing deep learning based automated coding algorithms and introduced a new architecture, Hierarchical Label-wise Attention Network (HLAN) and a label embedding initialisation approach for automated medical coding. We tested the approaches on the benchmark datasets extracted from the MIMIC-III database, with the simulated task to predict ICD-9 codes related to the high-risk diseases selected by the NHS for shielding during the COVID-19 pandemic. The experiment results showed that HLAN has a more comprehensive explainability and better or comparative results to the previous state-of-the-art, CNN-based approaches and the downgraded models, HAN and HA-GRU. The proposed label embedding initialisation effectively boosted the performance of the state-of-the-art deep learning models, capturing label correlations from the dataset, which reflects the coding practice.

Analyses on the experiment results of this work suggest that future studies are required in several areas: incorporating external knowledge, learning to capture missed coding, rare labels, and emerging new labels. In particular, automated medical coding work requires to be tested in real-world clinical settings and iteratively improved with inputs from relevant professionals such as coders, nurses and clinicians. Thus an open area to work on in the future is to adapt automated coding models with human corrections in real-time, which is mostly related to human-in-the-loop machine learning and active learning \cite{monarch2021}. Inspire by these, we plan to further test and develop the approach to support the coding department in the NHS. We will consult professionals to identify and address the issues involved in deploying the system to facilitate coding staff and to improve efficiency, accuracy, and overall satisfaction.

\section*{Acknowledgement}
The authors would like to thank Dr Johnson Alistair in the MIMIC-III team to confirm to display the sentences of discharge summaries in this paper. The authors would also like to thanks comments from Prof Cathierine Sudlow and other members in the Clinical Natural Language Processing Research Group in the University of Edinburgh. HD is supported by Health Data Research UK (HDR UK) National Phenomics Resource Project; Wellcome Institutional Translation Partnership Award (PIII032). VSP is supported by HDR UK National Text Analytics Implementation Project; Wellcome Institutional Translation Partnership Award (PIII029). HW is supported by HDR UK fellowship MR/S004149/1; Wellcome Institutional Translation Partnership Award (PIII054); The Advanced Care Research Centre Programme at the University of Edinburgh; The Health Foundation (I-qual-PPC). This work has also made use of the resources provided by the Edinburgh Compute and Data Facility (ECDF) (http://www.ecdf.ed.ac.uk/).

\bibliographystyle{elsarticle-num}
\bibliography{med_coding_ref}

\end{document}


\onecolumn
\pagenumbering{gobble}

\begin{table*}[t]
\caption{List of ICD-9 codes in \textbf{MIMIC-III-50} (50 codes, sorted by frequency in the training data) and per-label prediction results using Hierarchical Label-wise Attention Network with label embedding initialisation (HLAN+LE).}\label{code-50}
\center
\begin{threeparttable}
\begin{tabular}{p{2.5cm}p{4.5cm}|p{2cm}p{2cm}|lll}
\cline{1-7}
\textbf{MIMIC-III-50} ICD-9 code   & Short Title$^*$ & Frequency (train, 8066 documents) & Frequency (test, 1729 documents) & Precision & Recall & $F_1$ \\
\cline{1-7}
401.9  & Hypertension NOS         & 3233       & 778       & 78.5 & 82.8 & 80.5 \\
38.93  & Venous cath NEC          & 2139       & 402       & 60.6 & 47.9 & 53.4 \\
428.0  & CHF NOS                  & 2115       & 422       & 85.5 & 78.6 & 81.9 \\
427.31 & Atrial fibrillation      & 1992       & 470       & 91.7 & 88.6 & 90.1 \\
414.01 & Crnry athrscl natve vssl & 1921       & 435       & 84.4 & 72.0 & 77.7 \\
96.04  & Insert endotracheal tube & 1581       & 233       & 59.4 & 60.2 & 59.7 \\
96.6   & Entral infus nutrit sub  & 1525       & 228       & 67.0 & 63.4 & 65.1 \\
584.9  & Acute kidney failure NOS & 1448       & 362       & 71.4 & 56.4 & 62.9 \\
250.00 & DMII wo cmp nt st uncntr & 1416       & 340       & 73.2 & 69.5 & 71.3 \\
96.71  & Cont inv mec ven <96 hrs & 1395       & 258       & 64.2 & 49.5 & 55.7 \\
99.04  & Packed cell transfusion  & 1287       & 51        & 19.6 & 22.2 & 20.4 \\
272.4  & Hyperlipidemia NEC/NOS   & 1259       & 548       & 74.7 & 81.5 & 77.9 \\
518.81 & Acute respiratry failure & 1186       & 255       & 65.3 & 56.1 & 60.2 \\
39.61  & Extracorporeal circulat  & 1096       & 226       & 94.6 & 96.3 & 95.5 \\
599.0  & Urin tract infection NOS & 1067       & 251       & 73.8 & 59.2 & 65.7 \\
96.72  & Cont inv mec ven 96+ hrs & 969        & 187       & 66.5 & 61.1 & 63.4 \\
530.81 & Esophageal reflux        & 953        & 266       & 72.2 & 67.5 & 68.3 \\
272.0  & Pure hypercholesterolem  & 926        & 155       & 58.0 & 39.6 & 47.0 \\
285.9  & Anemia NOS               & 852        & 247       & 30.3 & 2.0  & 3.6  \\
88.56  & Coronar arteriogr-2 cath & 801        & 157       & 87.4 & 71.3 & 78.5 \\
38.91  & Arterial catheterization & 773        & 148       & 46.4 & 2.8  & 5.2  \\
486    & Pneumonia, organism NOS  & 765        & 174       & 63.7 & 51.0 & 56.1 \\
244.9  & Hypothyroidism NOS       & 761        & 210       & 80.8 & 84.0 & 82.3 \\
99.15  & Parent infus nutrit sub  & 736        & 91        & 82.2 & 62.3 & 70.9 \\
285.1  & Ac posthemorrhag anemia  & 726        & 203       & 69.4 & 44.8 & 54.3 \\
36.15  & 1 int mam-cor art bypass & 719        & 135       & 94.9 & 92.9 & 93.8 \\
276.2  & Acidosis                 & 694        & 195       & 72.6 & 25.4 & 37.6 \\
496    & Chr airway obstruct NEC  & 646        & 160       & 65.4 & 49.3 & 55.9 \\
995.92 & Severe sepsis            & 613        & 165       & 70.6 & 49.2 & 57.8 \\
V58.61 & Long-term use anticoagul & 604        & 181       & 66.3 & 66.9 & 66.1 \\
507.0  & Food/vomit pneumonitis   & 569        & 102       & 59.6 & 54.4 & 56.7 \\
038.9  & Septicemia NOS           & 567        & 149       & 52.5 & 28.1 & 36.1 \\
39.95  & Hemodialysis             & 549        & 72        & 84.4 & 80.4 & 82.3 \\
585.9  & Chronic kidney dis NOS   & 544        & 172       & 56.8 & 42.3 & 48.4 \\
88.72  & Dx ultrasound-heart      & 530        & 84        & 51.8 & 22.5 & 31.1 \\
410.71 & Subendo infarct, initial & 520        & 91        & 66.4 & 41.4 & 50.3 \\
403.90 & Hy kid NOS w cr kid I-IV & 513        & 215       & 77.5 & 62.9 & 69.2 \\
305.1  & Tobacco use disorder     & 504        & 181       & 28.3 & 4.6  & 7.2  \\
276.1  & Hyposmolality            & 494        & 155       & 52.7 & 26.6 & 34.9 \\
311    & Cutaneous mycobacteria   & 493        & 196       & 40.0 & 20.9 & 26.9 \\
37.22  & Left heart cardiac cath  & 482        & 100       & 62.4 & 44.7 & 51.9 \\
V45.81 & Aortocoronary bypass     & 479        & 108       & 76.3 & 67.3 & 71.5 \\
412    & Pneumococcus infect NOS  & 477        & 121       & 60.8 & 42.3 & 48.8 \\
287.5  & Thrombocytopenia NOS     & 471        & 138       & 58.3 & 23.3 & 33.0 \\
424.0  & Mitral valve disorder    & 451        & 104       & 72.9 & 38.3 & 49.8 \\
37.23  & Rt/left heart card cath  & 438        & 60        & 47.3 & 19.2 & 27.0 \\
511.9  & Pleural effusion NOS     & 421        & 95        & 46.5 & 16.8 & 24.4 \\
45.13  & Sm bowel endoscopy NEC   & 415        & 87        & 60.2 & 70.7 & 64.6 \\
33.24  & Closed bronchial biopsy  & 407        & 127       & 75.8 & 50.1 & 60.0 \\
V15.82 & History of tobacco use   & 397        & 187       & 0.0  & 0.0  & 0.0         \\
\cline{1-7}
\end{tabular}
\begin{tablenotes}
\item * Short titles of the ICD-9 codes are from \url{https://mimic.physionet.org/mimictables/d_icd_diagnoses/} and \url{https://mimic.physionet.org/mimictables/d_icd_procedures/}.
\end{tablenotes}
\end{threeparttable}
\end{table*}

\begin{table*}[th]
\caption{List of ICD-9 codes in \textbf{MIMIC-III-shielding} (20 codes, sorted by frequency in the training data) and per-label prediction results using Hierarchical Label-wise Attention Network with label embedding initialisation (HLAN+LE).}\label{code-shielding}
\center
\begin{threeparttable}
\begin{tabular}{p{2.5cm}p{4.5cm}|p{2cm}p{2cm}|lll}
\cline{1-7}
\textbf{MIMIC-III-shielding} ICD-9 code   & Short Title & Frequency (train, 4574 documents) & Frequency (test, 322 documents) & Precision & Recall & $F_1$ \\
\cline{1-7}
197.0  & Secondary malig neo lung                                                      & 656                         & 42                        & 82.1         & 84.8        & 83.3    \\
745.5  & Secundum atrial sept def                                                      & 592                         & 41                        & 95.3         & 83.7        & 89.0    \\
996.81 & Compl kidney transplant                                                       & 480                         & 14                        & 94.1         & 97.1        & 95.4    \\
042    & Shigella boydii                                                               & 470                         & 30                        & 97.7         & 99.3        & 98.5    \\
441.2  & Thoracic aortic aneurysm                                                      & 430                         & 36                        & 86.3         & 73.9        & 79.5    \\
416.0  & Prim pulm hypertension                                                        & 375                         & 10                        & 54.0         & 60.0        & 56.7    \\
746.4  & Cong aorta valv insuffic                                                      & 263                         & 35                        & 94.5         & 76.3        & 84.0    \\
288.00 & Neutropenia NOS                                                               & 196                         & 39                        & 51.2         & 18.5        & 26.5    \\
238.75 & Myelodysplastic synd NOS                                                      & 170                         & 21                        & 87.4         & 50.5        & 63.5    \\
996.82 & Compl liver transplant                                                        & 169                         & 4                         & 49.5         & 77.5        & 60.1    \\
238.71 & Essntial thrombocythemia                                                      & 164                         & 28                        & 74.6         & 46.8        & 56.4    \\
494.0  & Bronchiectas w/o ac exac                                                      & 152                         & 27                        & 87.6         & 71.1        & 76.5    \\
288.0  & Neutropenia                                                                   & 136                         & 0                         & 0.0          & 0.0         & 0.0     \\
996.85 & Compl marrow transplant                                                       & 133                         & 8                         & 71.9         & 77.5        & 74.3    \\
238.7  & Neoplasm of uncertain behavior of other lymphatic and hematopoietic   tissues & 116                         & 0                         & 0.0          & 0.0         & 0.0     \\
770.2  & NB interstit emphysema                                                        & 108                         & 0                         & 0.0          & 0.0         & 0.0     \\
501    & Alastrim                                                                      & 103                         & 5                         & 44.4         & 56.0        & 48.8    \\
288.03 & Drug induced neutropenia                                                      & 96                          & 17                        & 47.0         & 16.5        & 23.0    \\
289.59 & Spleen disease NEC                                                            & 95                          & 14                        & 71.9         & 45.0        & 55.0    \\
446.4  & Wegener's granulomatosis                                                      & 49                          & 1                         & 20.0         & 20.0        & 20.0    \\
\cline{1-7}
\end{tabular}
\begin{tablenotes}
\item * Short titles of the ICD-9 codes are from \url{https://mimic.physionet.org/mimictables/d_icd_diagnoses/} and \url{https://mimic.physionet.org/mimictables/d_icd_procedures/}.
\end{tablenotes}
\end{threeparttable}
\end{table*}

\begin{figure*}[th]
\caption{Distribution of label frequency in the training data for the datasets, MIMIC-III, MIMIC-III-50, and MIMIC-III-shielding.}\label{fig:dist-mimic}
\begin{center}
\includegraphics[width=0.6\textwidth]{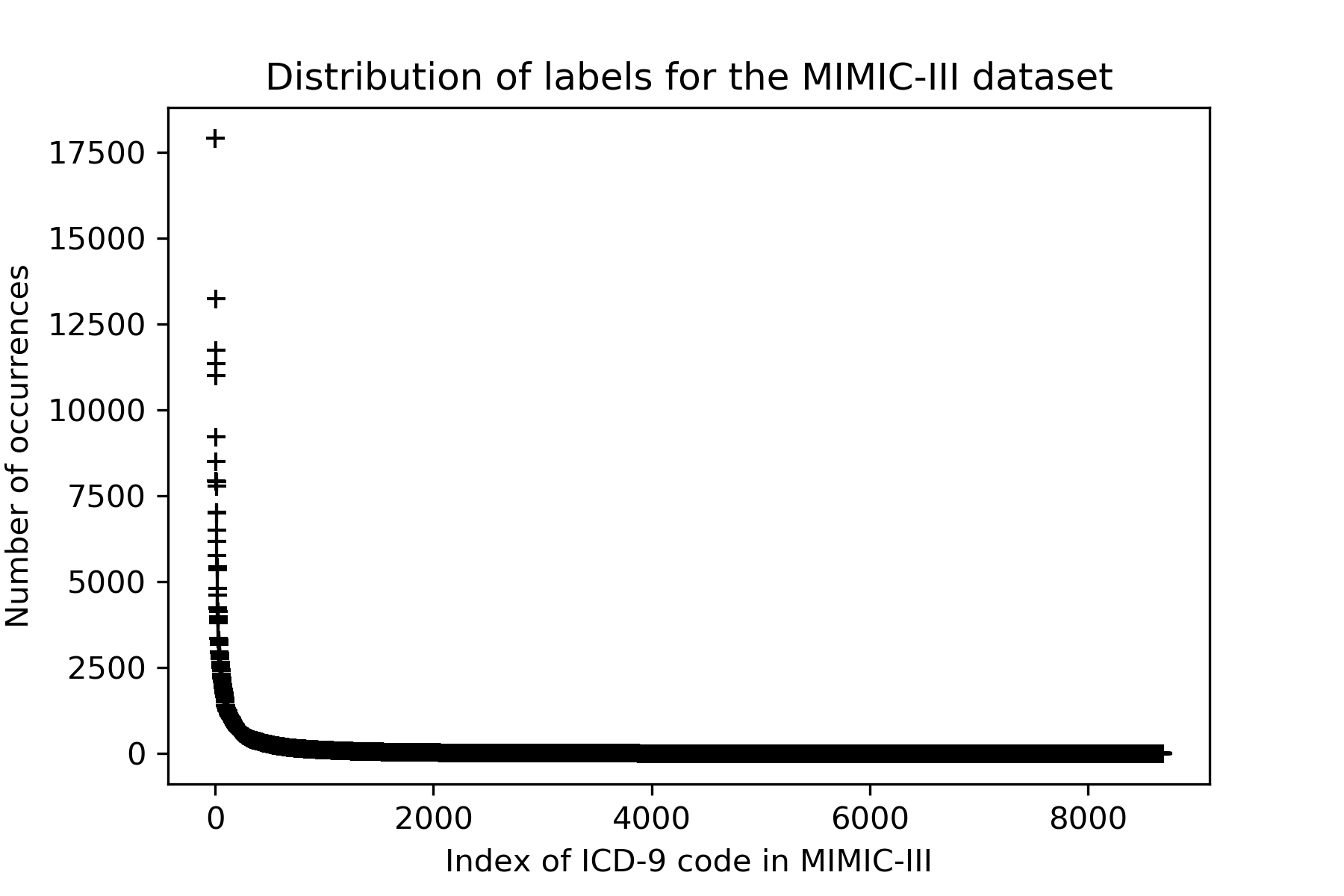}\\
\includegraphics[width=0.6\textwidth]{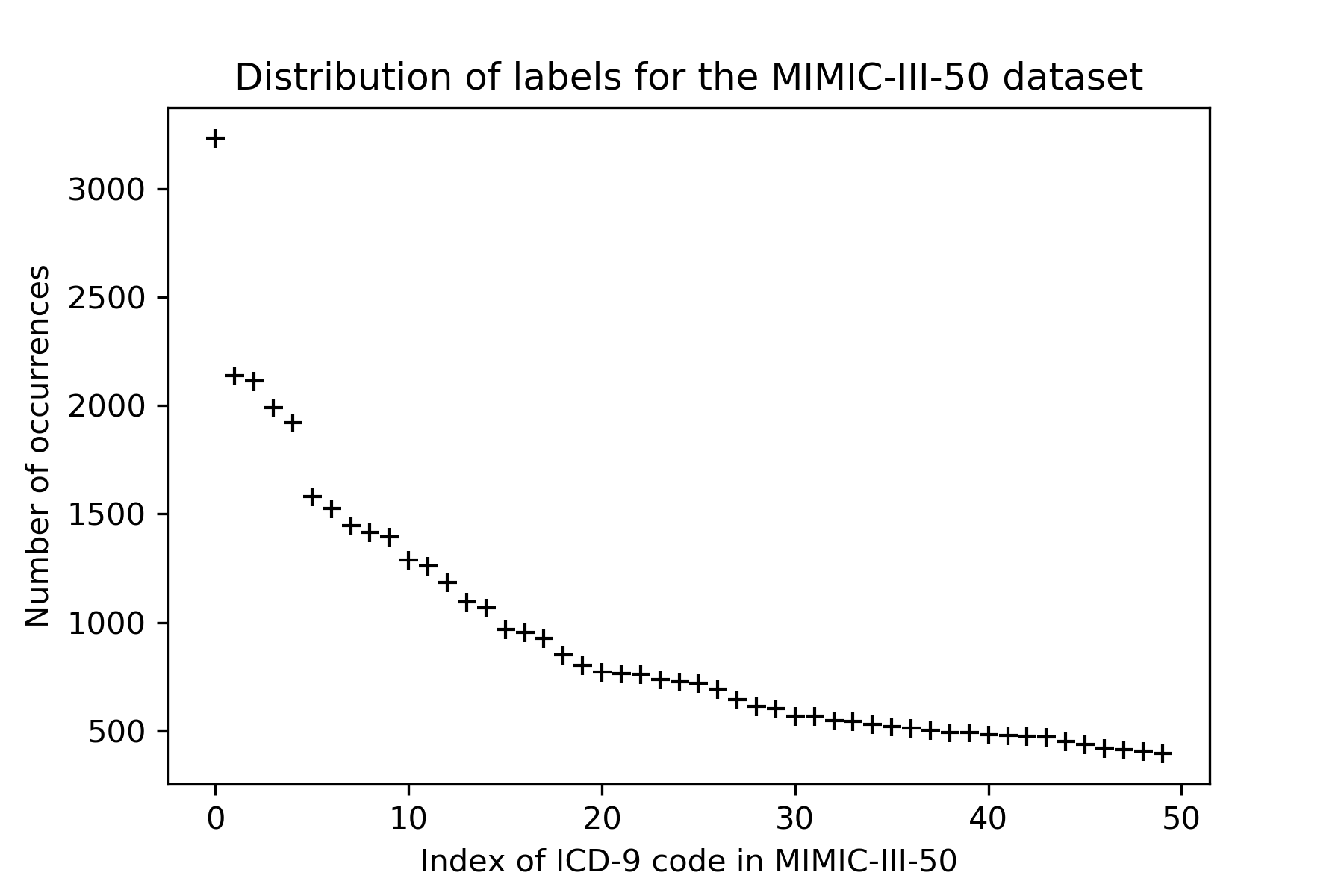}\\
\includegraphics[width=0.6\textwidth]{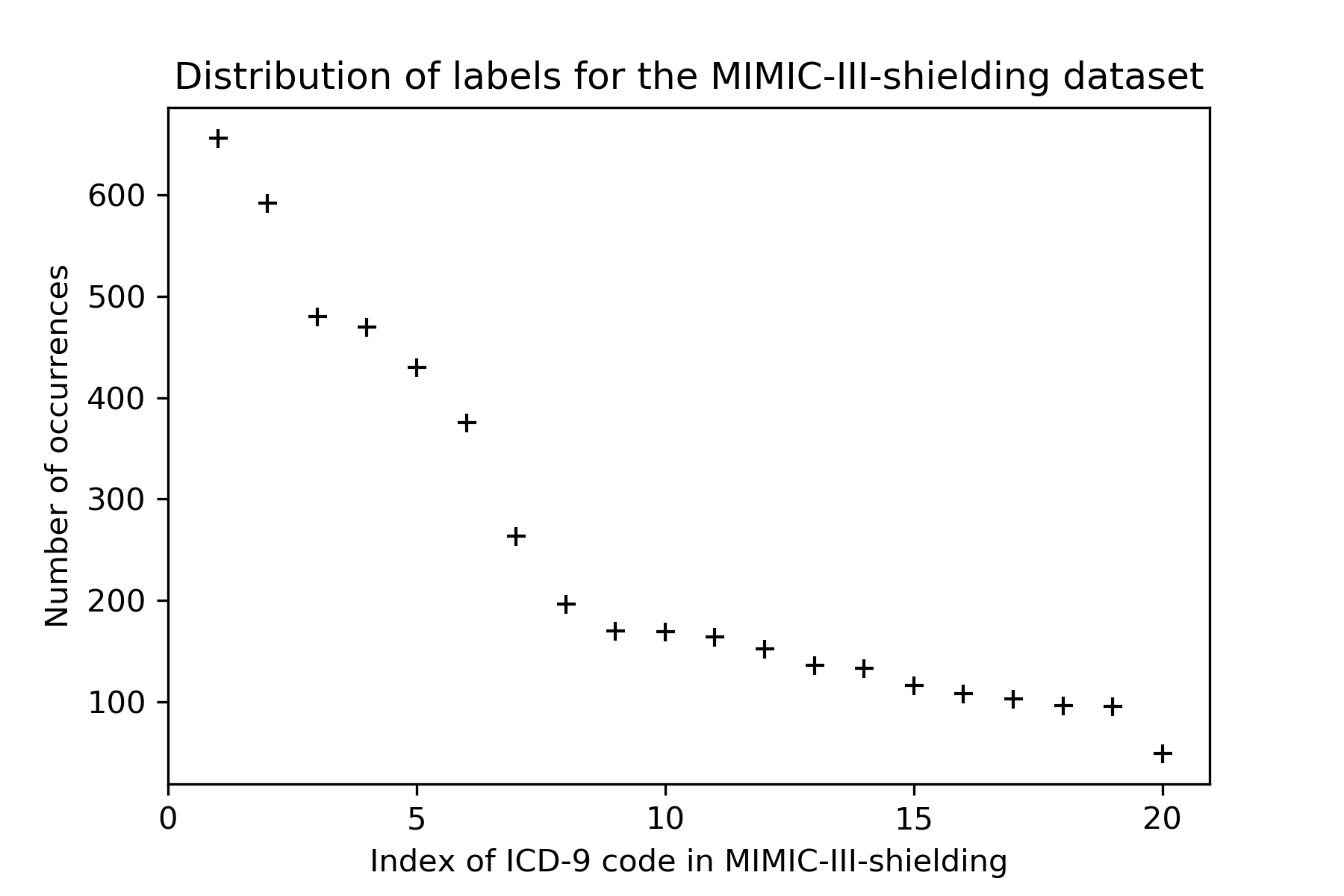}
\end{center}
\end{figure*}

\begin{table*}[th]
\caption{Model parameters, training time, and testing time from the datasets.}\label{model-setting-time}
\center
\begin{threeparttable}
\begin{tabular}{lllllll}
\cline{1-7}
                                                  & CNN*    & CNN+att* & Bi-GRU*  & HAN    & HA-GRU    & HLAN     \\
\cline{1-7}
\textbf{Parameter settings}\\
Calibration threshold $\mathit{Th}$                             & 0.5 & 0.5 & 0.5 & 0.5 & 0.5 & 0.5 \\
Learning rate                                     & 0.003  & 0.0001  & 0.003  & 0.01   & 0.01     & 0.01     \\
Batch size (training and testing)                 & 16     & 16      & 16     & 128    & 32       & 32       \\
Kernel size (or filter size)                                       & 4      & 10      & -      & -      & -        & -        \\
\# of words per document                               & 2500   & 2500    & 2500   & 2500   & 2500     & 2500     \\
\# of words per sentence $\mathrm{n_t}$           & -      & -       & -      & 25     & 25       & 25       \\
\# of sentences per document $n$                                & -      & -       & -      & 100    & 100      & 100      \\
\# of filters                                     & 500    & 50      & 512    & -   & -      & -      \\
Hidden size $d_h$                                 & -    & -      & -    & 100   & 100      & 100       \\
Attention layer size (e.g. $d_w$, $d_s$ in HLAN)                              & 500    & 50      & -      & 200    & 200      & 200    \\
Final hidden layer size                           & 500    & 50      & 512    & 400    & 400      & 400      \\
Dropout rate                                      & 0.2    & 0.2     & 0      & 0.5    & 0.5      & 0.5      \\
$L_2$ penalty                                        & 0      & 0       & 0      & 0.0001 & 0.0001   & 0.0001   \\
\cline{1-7}
\multicolumn{7}{l}{\textbf{Training time, estimated in \textit{minutes}$^{**}$}} \\
From MIMIC-III-50         & 5   & 50   & 40-50  & 10  & 30    & 80    \\
From MIMIC-III-shielding   & 2.5 & 8    & 20-40  & 10  & 10-15 & 25-30 \\
From MIMIC-III             & 250 & 1700 & 100-140 & 100 & -        & -        \\
\cline{1-7}
\multicolumn{7}{l}{\textbf{Testing time per document, estimated in \textit{milliseconds}, GPU time / CPU time}$^{***}$} \\
From MIMIC-III-50         & 2 & 5 & 50 & 34 / 40 & 61 / 160 & 141 / 330\\
From MIMIC-III-shielding   & 3 & 3 & 43 & 32 / 40 & 17 / 30 & 14 / 50 \\
From MIMIC-III            & 2 & 3 & 50 & 42 / 40 & - & -\\
\cline{1-7}
\end{tabular}
\begin{tablenotes}
\item ``-'' denotes that the parameter is inapplicable to the model or the estimated time was not obtained.
\item * Parameter settings for CNN, CNN+att, and Bi-GRU are the same as in Mullenbach et al., 2018.
\item ** All models were trained and tested using a single GeForce GTX TITAN X server.
\item *** For HAN, HA-GRU, and HLAN, testing times on a CPU server (4-core, Intel(R) Xeon(R) Platinum 8259CL CPU @ 2.50GHz) were further reported (displayed after the GPU time).
\end{tablenotes}
\end{threeparttable}
\end{table*}